\documentclass[journal]{IEEEtran}
\usepackage{graphicx}
\usepackage{subcaption}
\usepackage{hyperref}
\usepackage{amsmath}
\usepackage{algorithm}
\usepackage[noend]{algpseudocode}
\usepackage{multirow}
\usepackage{array}
\usepackage{xcolor}
\newcommand{\sas}{S\&S} %Subject and Subjectivity Shortform
\newcommand{\change}[1]{\textcolor{black}{#1}}
\newcommand{\newchange}[1]{\textcolor{black}{#1}}
\begin{document}

\title{Tension Space Analysis for Emergent Narrative}
\author{Ben~Kybartas,
        Clark~Verbrugge,
        and~Jonathan~Lessard
\thanks{B. Kybartas and C. Verbrugge are with the School of Computer Science of McGill University, Montr\'{e}al, QC, Canada}
\thanks{J. Lessard is with Concordia University, Montr\'{e}al, QC, Canada.}}

\markboth{Transactions on Games}%
{Kybartas \MakeLowercase{\textit{et al.}}: Tension Space Analysis for Emergent Narrative}

\IEEEoverridecommandlockouts
\IEEEpubid{\makebox[\columnwidth]{~\copyright2020 IEEE \hfill} \hspace{\columnsep}\makebox[\columnwidth]{ }}
\maketitle
\IEEEpubidadjcol

\begin{abstract}

Emergent narratives provide a unique and compelling approach to interactive storytelling through simulation, and have applications in games, narrative generation, and virtual agents. However the inherent complexity of simulation makes understanding the expressive potential of emergent narratives difficult, particularly at the design phase of development. In this paper, we present a novel approach to emergent narrative using the narratological theory of \textit{possible worlds} and demonstrate how the design of works in such a system can be understood through a formal means of analysis inspired by \textit{expressive range analysis}. Lastly, we propose a novel way through which content may be authored for the emergent narrative system using a sketch-based interface.

\end{abstract}

\begin{IEEEkeywords}
Interactive Narrative, Interactive Storytelling, Emergent Narrative, Character Modelling, Conflict Modelling
\end{IEEEkeywords}

\section{Introduction}
\label{sec:intro}

\IEEEPARstart{E}{mergent} narratives (EN), an approach to narrative in which stories are created through a simulation of characters in a virtual storyworld, drive some of the most interesting interactive storytelling works available today, from academic experiments~\cite{McCoy2014, aylett2005} and independent games~\cite{adamsDwarfFortress2006, cavesofqud} to successful commercial franchises~\cite{paradoxCrusaderKings22012, maxisSims}. Such works make less use of typical models of narrative, instead relying on the emergence of complex, narratively compelling sequences of character behaviour that occur over the duration of the simulation, typically in response to interaction from a human interactor. As proposed by Ryan, emergent narratives create a form of \textit{``nonfiction''}, in the sense that these interesting events actually occurred through the natural progression of the simulation, rather than being authored or explicitly generated~\cite{jryanphd}. Eladhari further states that the ability to discover event sequences worth re-telling is a crucial measure of the success of an EN work~\cite{eladhari2018}.

Nonetheless the creation of works in EN is challenging quite simply through its complexity, both in the \textit{systems} that drive the simulation, and the \textit{content} which the system is provided. Essentially, an EN work must be provided with a large body of content, which controls the behaviour of the systems underneath. Both have a large impact on the type of stories and events which can be produced. Systemically, EN works often rely on a large number of interacting complex systems, and these interactions grow combinatorially for each new system. The content of an EN work typically embeds the work within a context, either a genre of narrative or storyworld that is ``narratively pregnant'' with a number of possible stories~\cite{mateasInteractiveDramaPhD2002}. The problem, however, is that the impact that content will have upon the range of possible stories is not clearly defined, hindering the ability to author for particular outcomes.

In EN works, the systems and content are often developed separately, particularly when separate individuals or teams are responsible for each. We argue, however, that given the complexity of EN, any meaningful form of authorship must also embed an \textit{understanding} of the narrative potential of each piece of content within the greater system. Tools for smart authorship and analysis of interactive storytelling work have been explored in academic works, often focused on the evaluation of existing authored content~\cite{garbeIceBound2014}, or through analyzing the choices made by interactors in a complete EN work~\cite{McCoy2014}. Such approaches are extremely valuable, and we argue that this knowledge could be taken further, and essentially allow the creation of content \textit{through} its intent in the EN work. This approach is echoed in a number of declarative modelling~\cite{smelikDeclarativeModeling2011} and mixed-initiative design tools~\cite{liapis2013sketchbook, compton-iccc2015}. \newchange{Nonetheless, a number of limitations bottleneck the widespread usage of such design tools, such as their limited availability and lack of tools for narrative content analysis~\cite{koenitzchallenges2019}, and supports the view that developing authoring tools should also be embedded with tools for content analysis.}
%Thus, development of new tools for authoring should be coupled with progress in the creation of new means by which EN works may be analyzed, independent of the tool itself.}

In this paper we focus on the development of techniques for analysis and content authorship for a novel EN system, based upon the narratological theory of possible worlds. A proof-of-concept version of the system was previously described and used in a released EN work~\cite{kybartas_2017, subjectandsubjectivity}. Here, we formalize the model and properties of the system at a foundational level. We further present a procedure for analyzing the content of the EN system using techniques similar to those of~\textit{expressive range analysis}. This technique was previously explored at a theoretical level~\cite{kybartas2018}, but is here formalized and further realized in the form of an authoring tool. Lastly, we propose a novel technique for content authorship, in which the same techniques used for the analysis of EN works can be used to author content through the use of a sketch-based interface, which is also realized in the form of an authoring tool.

Formally, the four main contributions of this paper are as follows:
\begin{itemize}
	\item A formal model and implementation of a character-focused emergent narrative grounded in the narratological theory of possible worlds.
	\item A set of metrics for analyzing both the content and traces of a given EN work, utilizing a novel approach based upon expressive range analysis.
	\item A conceptual sketch-based approach to authoring content for EN works.
	\item A proof-of-concept authoring tool capable of authoring content, performing analysis, and sketch-based authoring.
\end{itemize}

As a note on terminology, in this paper we use \textit{EN system}, to refer to the systemic side of the emergent narrative. When we refer to \textit{content}, we refer specifically to content which actively affects the behaviour of the system, ignoring for example 3D models and textures. We use \textit{EN work} to refer to a final product made within the EN system. We adopt the term \textit{interactor} to describe a human participant. We use \textit{trace} to refer to a particular sequence of a given simulation.

\section{Background}

In this section we briefly present background work on possible worlds  theory, upon which the system is based, and on \change{expressive range analysis}. Though non-exhaustive, the core theories are presented to help understand the remainder of the paper.

\subsection{Possible Worlds Theory}
\label{sec:pwbackground}
The possible worlds theory arose from discussions of the nature of truth and believability in fiction, and is centered around the concept that a fictional narrative occurs in a  fictional world, that can be treated as a \newchange{\textit{possible world} (PW)} that exists alongside our actual world~\cite{Pavel1975}. The \newchange{PW} differs from the real world in both its facts and internal workings, and when reading a reader will ``recenter'' and align their perspective with the world of the narrative, filling in missing information with their understanding of the real world~\cite{dolozel2000}. This allows, for example, a reader to accept a world with fantastical elements such as magic, while still expecting the world to obey certain natural laws.

In this article, we concern ourselves principally with work by Ryan~\cite{Ryan1991}, who applied \newchange{PW}s theory to characters. Ryan treats a narrative as a large model structure comprising of a number of \newchange{PW}s, at the center of which is the ``actual'' \newchange{PW} of the narrative. The remaining \newchange{PW}s define the characters, where specific \newchange{PW}s define a specific component of the character, such as their \textit{knowledge}, \textit{obligations}, \textit{intents} and \textit{wishes}. The knowledge world represents what the character believes to be true about the actual \newchange{PW}, and can even be extended to contain an entire sub-structure of the narrative (containing what the character believes the other characters believe and so forth). Obligation and wish-worlds, are treated as \textit{ideals}, in that each represents a possible state that of the actual world that the character wants to be true, or feels obliged to be true. Ryan proposes that all actions the characters take in a narrative, are done to attempt to shift the state of the actual world so that it matches all of their obligation and wish-worlds, with plans being stored in intent PWs. Since these different ideals conflict, either internally within a character or interpersonally between characters, the actual PW can never match every one of the ideals, and the discovery and resolution of these conflicts forms the driving force of the narrative.

Ryan's model is a compelling candidate to explore for EN in that it presents a model of narrative and character that is not focused on the actual plot, but rather represents a potential for a number of possible plots. Its narratological focus is also an interesting alternative to the psychological or sociological models of character more commonly found in EN works~\cite{McCoy2014, maxisSims, evansVersu2014}. Further, its subjective model of characters, with varying ideals and knowledge, supports the development of a number of rich narrative features, as outlined by a number of different authors~\cite{Lessard2016, harrellbook2013, sgourosQuantumConceptsNarrative2015}.

\subsection{Expressive Range Analysis}
\label{sec:conflictbackground}

Expressive range analysis (ERA), \change{was introduced by Smith and Whitehead and} has its origins in the field of generative systems and procedural content generation~\cite{smithExpressiveRange2010}. In this field, ERA consists of defining metrics according to which generated content can be evaluated, and then plotting the evaluation of a number of generated examples using a heatmap. This gives a sense of what ``range'' of possibilities are possible given the current set up of the generator. This can then be used to fine-tune the parameters of the generative methods, or to experiment with different content.

Since in this paper we look at authorship at the foundational level, and further are more interested in the presence of narrative properties and system behaviour as opposed to qualitative analysis, \newchange{ERA is a good fit for analysis~\cite{kybartas2018}. The simple visualization and ability to compare ranges dynamically as content is changed make it a useful addition for content creation tools.}

Defining the EN system as a generative method or content generator is slightly more challenging. While it makes sense to treat the EN as a narrative generator, what comprises a ``good'' narrative in an EN work is often significantly unclear, and further may vary according to the intent of the author for the work. We opt instead to treat the EN as a \textit{conflict} generator, since conflicts are clearly defined in Ryan's PW work, are clearly acknowledged as a foundational component to narrative, and further have been formalized and explored in interactive storytelling research~\cite{wareConflict2014, szilasDrama2017}. This can be thought of theoretically as a formal means of evaluating the ``narrative pregnancy''~\cite{mateasInteractiveDramaPhD2002} of a space according to conflict, focused more on the ways narrative properties will occur independent of any evaluation of the the resulting narratives.

\section{Related Work}

The field of interactive storytelling (IS) in which EN is based is large and diverse~\cite{7439785}, and many key works lie outside the scope of this paper. As such, in this section we focus specifically on several specific fields and works in IS and the broader research community that are closely related to this work, namely in emergent narratives, character modelling for IS, and formal analysis for authoring tools.

\subsection{Emergent Narrative}

The term emergent narrative, in the context of the IS community, is often linked back to an early work by Aylett~\cite{aylett1999}. In this work, Aylett presents the well known challenge in IS of maintaining an author's narrative intent while still allowing an interactor to meaningfully impact the narrative, proposing that this intent instead be embedded within the models of characters used in a simulated world. Aylett further makes connections between emergent narrative, theatre and improvisation, which remains an area of interest for interactive storytelling practitioners~\cite{swartjesLateCommitment2008, Fuller2010}. Aylett et al.'s later well known \textit{FearNot!} project~\cite{aylett2005}, is an emergent narrative geared towards children's education regarding bullying. It involves the simulation of a number of characters, most importantly a ``bully'' and ``victim'' character, with children adopting the role of the victim character and being able to explore a number of coping mechanisms to the bully's harassment. \change{We adopt Aylett's description of emergent narrative when we refer to our own system, however our system differs from \textit{FearNot!} in that we focus on a multiplicity of ways conflict may occur based upon Ryan's possible worlds theory.} 

Both McCoy et al.~\cite{McCoy2014} and Evans and Short~\cite{evansVersu2014} explored the creation of EN works through highly complex and detailed social models. For McCoy et al., their EN work \textit{Prom Week} was realized through an elaborate model of social interaction, melding deep character models with models of normative and cultural behaviour. In Evans and Short's Versu system, a model of \textit{social practices} created a set of rules and actions which limited character behaviour, that was still driven by a set of character desires and beliefs. Both systems had rich character behaviours, but involved a significant amount of authorship, as well as more practical concerns such as how to convey character's decision making processes. \change{Both systems involve a significantly more detailed social simulation than the system we present here, but both systems noted high difficulty with authorship, and addressing these concerns is one of the goals of our research.}

Our system does not make use of psychological or sociological models for character, instead opting for a narratological approach that is arguably founded upon character \textit{values} as realized through a set of ideals. Szilas' \textit{Nothing for Dinner}~\cite{nothingfordinner2014} and its underlying \textit{IDTension} system~\cite{szilasIDTension2003fixed} explicitly defines its characters according to a set of values, and the work is focused on the conflicts created from the overall system of values. \change{Szilas' approach to creating tension is similar to our own, however we use a different model of values, where Szilas treats values as something that may or may not hinder the separate goals of a character, we treat values \textit{as} the goals of the character.}
%Although not a traditional emergent narrative, it is worth mentioning that Mateas and Stern's \textit{Fa\c{c}ade} also approached the development and exploration of two character's conflicts of values~\cite{mateas2003faccade}, but opted for an authoring heavy and drama-management approach rather than character simulation. The result is a set of two richly defined characters, but with the richness mostly arising from the authored elements and drama-management rather than through models and simulation.

\subsection{Character Modelling}

As with our own system, a number of academic works have explored the modelling of modal and epistemic properties of a characters. Both Thorne and Young~\cite{AIIDE1715844} and Shirvani et. al.~\cite{Shirvani_Ware_Farrell_2017} have explored the impact of beliefs, particularly false beliefs, in the context of planning for plot generation. With varying and potentially incorrect beliefs, the planner is able to represent forms of character surprise, manipulation and deception.  \change{\textit{Thespian}, a multi-agent interactive narrative uses the ```theory of mind'' approach to character modelling, where characters similarly reason about each others beliefs, and personality is modelled in how each character weights their respective goals~\cite{Si:2005:TUM:1082473.1082477}. Interestingly, characters may automatically be fit by extracting goals and beliefs from a manually authored linear ``scripts'' of the desired behaviour of each character. Work from Alfonso et al. also uses a theory of mind approach, but extends their work to incorporate reasoning over the potential emotions and emotional behaviours of other agents in the system~\cite{alfonso2015emotional}. Similarly, \textit{Mirage} is an interactive narrative architecture where characters are provided emotional and belief models, but can further make similar make such inferences about the human user who is interacting in the work~\cite{El-Nasr:2004:UAS:1067343.1067356}.} \change{At present, our system does not have character's reasoning about other character's internal mental model, however in each of these works, like our own, narrative effects are achieved by having explicit subjective worldviews for each character, that allow for narrative effects.}

While not specifically IS focused, Eger and Martens~\cite{egerBeliefManipulation} present a character knowledge model based upon modal logic and Baltag's action language, specifically exploring belief manipulation as a mechanic in social deduction games. Finally in work from Robertson and Young~\cite{robertson2019} and Swartjes et al.~\cite{swartjesLateCommitment2008}, the knowledge of the player is also tracked, allowing essentially ``behind the scenes'' manipulation of the game world to allow characters to create dramatic effects or mediate player actions that deviate from the intended story. \change{These approaches are based on having a belief about the state of the real world which may true or false, which is handy for creating a number of interesting narratives, and currently exists in our system although extending the analysis we provide in this paper to elements of belief is ongoing work.}

Our work, while modelling elements of belief and knowledge, instead treats the representation and interaction of character worldviews as the core of the EN system. Models of character worldviews have been used in the Chimeria~\cite{Harrell2018}, and  GRIOT~\cite{harrell2007b} systems from Harrell and colleagues, and also in Sgouros' work on modelling quantum mechanics in narrative~\cite{sgourosQuantumConceptsNarrative2015}. In each case characters have a subjective worldview or valuation that manipulates how they perceive and evaluate the world, in particular how these could model cultural beliefs or identities. Lessard and Arsenault, not approaching the modelling of worldviews directly, nonetheless outline a philosophy for interactive storytelling in which characters are treated as a ``subjective interface'' to the underlying model, able to warp and modify facts according to internal desires possibly unknown to the interactor or other characters~\cite{Lessard2016}. Harrell similarly outlines approaches to narrative works which intend to expose or explore underlying subjective worldviews as one of their artistic intents~\cite{harrellbook2013}. \change{Our system exists in a similar vein to these works, though structurally different, in that worldviews are considered the main drive of narrative and further that the subjectivity of the worldview is an important driver for interesting character behaviours and storytelling.}

Structurally, our system holds several similarities with the \textit{belief-desire-intention} (BDI) model, common in multi-agent systems, which has been explored for its ability to be used for character models in IS~\cite{peinado2008, berov2017}. \change{BDI was cited as an inspiration in Ryan's work, and the different types of worlds in the model (Eg. knowledge, wish, etc.) definitely have theoretical overlap with the BDI model, though explicitly applied to the types of dramatic characters since in narratives.} Berov, in particular, has also explicitly cited Ryan's work as the inspiration for a theoretical BDI framework which also includes emotion and personality~\cite{berov2017}. Although similar, our system does not model explicit intents, opting instead for selecting actions based on their immediate impact, more similar to the \textit{utility-based AI} approach taken in McCoy et al.'s \textit{Prom Week}~\cite{McCoy2014}.

\subsection{Formal Metrics and Authoring Tools}

Explorations of metric analysis techniques for IS works are typically geared to evaluating or discovering certain narrative properties within the work, independently from the qualitative evaluations that are typically performed through user studies. Though some research has aimed to quantify qualitative properties~\cite{kybartasReGEN2014}, most focus on the formalization and detection and/or quantification of certain narrative properties such as surprise~\cite{baesurprise2014}, the aforementioned tension~\cite{szilas2018}, and conflict~\cite{wareConflict2014}. \change{Notably, work from Partlan et. al utilized expressive range analysis to analyze strucutural properties of interactive, graph-based narratives~\cite{AIIDE1818103}.} In the same work which presented the possible worlds model, Ryan also introduced the concept of ``tellability''~\cite{ryanPossibleWorldsNarrative1991}, which aims to express the degree to which a narrative is interesting to ``tell'' to someone else, and emphasizes that the possible narratives that could occur are equally as important to making interesting narratives. Berov explores a formalization of this metric, applied to a BDI model of characters~\cite{berovTellability2017}.% J. Ryan also presents Ryan's tellability as a possible way to ``curate'' a given emergent narrative trace to discover potentially interesting stories~\cite{jryanphd}. 

Analysis techniques may also be used for debugging or refinement of IS works. Garbe et al. developed a technique for evaluating the authored content in their work \textit{Ice-Bound}, by exhaustively checking all possible combinations of input to discover problems and detect regions which need further authoring~\cite{garbeIceBound2014}. Si et al. also designed an authoring tool for creating characters within the \textit{Thespian} system, in which the author defines constraints on the desired character behaviour and goals, and the system then simulates a number of possible traces of the system, detecting any behaviour which deviates from the authorial intent~\cite{siProactiveAuthoring2007}.

In this paper we are interested in how this type of understanding can be extended to allow for the \textit{declarative} authoring of characters for EN works. \change{Declarative modelling describes an approach to authorship in which the author declares their intended desires for a piece of content at a level that is more abstract than the actual content which is required by the system, which are then realized into a piece of content through generative methods, allowing for fast content creation while respecting authorial intent~\cite{smelikDeclarativeModeling2011}.} This approach has been applied in various areas, such as using sketching tools to create virtual cities and terrains~\cite{smelikDeclarativeModeling2011} or video game levels~\cite{liapis2013sketchbook}. Similarly, Compton and Mateas' ``casual creators'', use declarative methods to make authoring accessible to non-experts, and approach content creation from a playful and exploratory angle~\cite{compton-iccc2015}.

\section{The Foundational Model}
\label{sec:model}

In this section we present the formalism, metrics and functionality of the foundation of our EN system. We use the term ``foundational'' to indicate that the system described here forms the core of the EN system that is fully functional, and can support any additional systems built on top of it (dialogue, emotion, memory, etc.). \newchange{Although here we focus solely on definition, a working example of this model will provided and analyzed in Section~\ref{sec:analysis}.} 

\subsection{Formalism}
\label{sec:formalism}

The \textbf{narrative system} is defined as the tuple $N = \langle P, T, w_a, C, A, r \rangle$, where $P$ is a set of \textit{propositions}, $T$ is a set of \textit{themes}, $w_a$ is the \textit{actual world},  $C$ is a set of \textit{characters}, $A$ is a set of \textit{actions}, and $r$ is a set containing both a  non-zero positive integer range of \textit{truth values} and a \textit{\change{``don't care''}} value, $\bot$.

A \textbf{proposition} may be thought of as a fact which can evaluate to any one of the integer range values in a given world. For example, in a basic true/false range ($r = [0, 1]$) a proposition may be ``it is raining'' which can either be true or false, whereas for a large range a proposition might be ``rain intensity'' which is graded according to the values in $r$.

A \textbf{theme} is a label that defines a perspective according to which characters will have an ideal world. In Ryan's work, there are two types of worlds which represent ideals, \textit{wishes} and \textit{obligations}. Here we adopt a more specific labelling, so a theme may be something like \textit{politics} for a political wish world or \textit{family} for a family obligation world. The ambiguous term ``theme'' was chosen both to avoid any restrictive classification, and since theme often refers to the driving force or underlying substance of a narrative work, which can be said as true for the system as presented.

The main content of the system is worlds, in that even characters are in essence a collection of different worlds. A \textbf{world} is a set of \textbf{truth values}, $V$, in the range of $r$, with one truth value for each proposition in $P$. We currently define two different types of worlds, an \textit{ideal} world representing the desirable state of the world and an \textit{epistemic} world which relates to either the actual state of, or a belief of the actual state of, the storyworld. Both types of worlds are structurally the same, except that an actual world is restricted from using the \change{``don't care''} ($\bot$) value, since it is necessary for each proposition to have a truth value. A world is, in essence an assignment of truth values to each one of the propositions. So, for example, if we have the proposition ``it is raining'', an actual world will assign whether this proposition is true or false according to the weather, whereas an idealized world will state whether the character wishes this proposition to be true or false, or does not care either way.

The \textbf{actual world}, $w_a$, represents the true state of the world inhabited by the characters in the narrative system. The actual world is both what is perceived by each character and what is influenced through their actions.

A \textbf{character}, $c$, is defined with the following tuple $c = \langle w_{p}, W_{v}, A_c \rangle$, where $w_p$ is the character's \textit{perceived actual world}, $W_v$ is a set of \textit{worldviews} and $A_c$ is the subset of actions $A$ available to the character. The \textbf{perceived actual world} is an epistemic world that represents what a character believes to be the state of the actual world, \change{which} allows characters to hold false beliefs regarding the state of the actual world. %, although at this stage we do not provide any significant functionality for knowledge and belief.
A \textbf{worldview} is an ideal world which is assigned to one of the available themes. In the system, each character must have one ideal world for each theme. Generally, a character's worldview is treated as \textit{fixed} and their ideals do not change throughout the work. As will be explained further in Section~\ref{sec:properties}, fixing the worldviews allows a consistent level of conflict in the narrative, although it is also fairly trivial to extend the system if shifting worldviews are required. The \textbf{set of actions} is a subset of the actions $A$ that are defined in $N$, and is the set of actions that a character may take in the system. A character's ability to affect the actual world, and therefore reach their ideals, is limited by the set of actions available to them.

An \textbf{action}, $a$, is defined by the following tuple $a = \langle w_{c}, w_{e} \rangle$, where $w_{\delta}$ is an epistemic world describing the action's \textit{precondition}, and $w_{e}$ is an epistemic world describing the action's \textit{postcondition}, or effect on the world. A \textbf{precondition} is a world where each value must match the value of the actual world in order for the action to be taken successfully, with the \change{``don't care''} operator being used when the action does not depend on the truth value of a specific proposition. The \textbf{postcondition} is a world which defines the new truth values of the actual world after the action is taken. For a postcondition, the \change{``don't care''} operator means the truth value of that particular proposition is not affected by the action.

\subsection{Properties}
\label{sec:properties}

Given the model of the system, we now define the properties which guide how \change{characters} behave and interpret the world. In this paper, we adopt the definitions of \textit{conflict} and \textit{tension} from Ware et al.~\cite{wareConflict2014} and Szilas~\cite{szilas2018}. Conflict, only occurs during the functioning of the system, and occurs at points where the intent of a character is challenged and possibly subverted by events in the story. Tension, is the \textit{potential} for conflicts to occur and can be evaluated before any narrative is actually generated. Ware et al. and Szilas' character models both have defined intents, but in our system, we focus specifically on Ryan's proposal that character's want all their worldviews to align with the actual world, which we call the \textit{overarching goal} of each character.

It is, however, generally impossible for any of the characters to reach this goal. Since each character must satisfy all their ideal worlds, if there are any differences between the truth values of ideal worlds, then no possible state of the actual world could match both worlds at once. We treat this as tension, since it is precisely at these points that conflict may occur. In our model, calculating the tension between two worlds is handled by taking the distance between the truth values of two worlds, ignoring \change{``don't care''} values. This is, in essence, the Manhattan distance between two worlds. To formally define distance, we first define the distance for any two truth values, $v_1$ and $v_2$ as follows:

\begin{equation}
\label{eq:vdist}
\textit{dist}(v_1, v_2) = 
	\begin{cases}
	0 & \quad \text{if }v_1 = \bot \vee v_2 = \bot\\
	|v_1 - v_2| & \quad \text{otherwise}
	\end{cases}
\end{equation}

From this, we derive our distance between worlds $w_1$ and $w_2$, as the summation of the distances between each pair of truth values:

\begin{equation}
\label{eq:wdist}
\textit{Distance}(w_1, w_2) = \sum_{i=0}^{|P|} \textit{dist}(w_1.v_i, w_2.v_i)
\end{equation}

Most of our metrics will be based upon this distance function. From this function, it is clear that any case where the distance is zero means the two worlds must either match, or only differ in cases where one value is a \change{``don't care''}. This property allows us to formalize our concept of the overarching goal, and state that a character's goal is that the \textit{total distance between each worldview and the actual world is zero}. Since the distance gives a measure of how ``far''  worlds are from one another, we define the \textbf{goal tension} as the sum of the distances of each worldview to the actual world:

\begin{equation}
\label{eq:gtens}
\textit{Tension}_{g}(c) = \sum_{i=0}^{|W_v|} \textit{Distance}(c.w_{v_i}, w_a)
\end{equation}

We can additionally define a secondary \textbf{subjective goal tension}, which is what the character \textit{believes} their goal tension to be, by instead comparing with the character's perceived actual world:

\begin{equation}
\label{eq:gstens}
\textit{Tension}_{\textit{gs}}(c) = \sum_{i=0}^{|W_v|} \textit{Distance}(c.w_{v_i}, c.w_p)
\end{equation}

We identify two further forms of tension which exist within the system, which are the tensions which exist \textit{between} worldviews, defined as the \textbf{personal tension} when referring to two worldviews of a single character, and \textbf{interpersonal tension} when referring to the worldviews of two differing characters. Formally, we first define the personal tension for a character $c$ and their two worldviews $c.w_{v_1}$ and $c.w_{v_2}$ as follows:

\begin{equation}
\label{eq:ptens}
\textit{Tension}_{\textit{p}}(c.w_{v_1}, c.w_{v_2}) = \textit{Distance}(c.w_{v_1}, c.w_{v_2})
\end{equation}

and similarly for interpersonal tension with characters $c_1$ and $c_2$ and worldviews $c_1.w_v$ and $c_2.w_v$:

\begin{equation}
\label{eq:itens}
\textit{Tension}_{\textit{i}}(c_1.w_v, c_2.w_v) = \textit{Distance}(c_1.w_v, c_2.w_v)
\end{equation}

Goal tension and interpersonal/internal tensions are both different in how they are understood by the system. The goal tension is dynamic, and can change with each action taken during simulation, whereas the worldview tensions are static, since the foundational system assumes fixed worldviews. The worldview tensions are thus the most important to consider when authoring, as it is here that the author defines the tensions that can eventually lead to conflict. \change{The goal tension is a slightly different form of tension, measuring the tension felt by the distance of the character to their overarching goal, where the potential for conflict here may occur due to any changes to the actual world that move the character further from this goal.}

There are also a number of base properties which can be extracted from the foundational model that are not included here, as they do not factor into the base functioning of the system, but include measures such as the correctness of an epistemic world (distance to the actual world), total tension (sum of all worldview tensions), normalized tension (percentage of total possible tension given size of the world), etc. Each of these has uses in possible extensions to the system.

\begin{algorithm}[H]
\caption{Base Functionality}
\label{alg:turn}
\begin{algorithmic}[1]
\Function{Step}{Narrative $N$}
\For{\textbf{each} $c \in N.C$}
\State $a_c \gets $ Act($c$)
\If{$\neg a_c = null$ \textbf{and} $\textit{Distance}(a_c,N.w_a) = 0$}
\State $N.w_a \gets $ Apply($a_c.w_\epsilon$, $N.w_a$)
\For{\textbf{each} $c \in N.C$}
\State $c.w_p \gets $Apply($a_c.w_\epsilon$, $c.w_p$)
\EndFor
\EndIf
\EndFor
\EndFunction
\end{algorithmic}
\end{algorithm}

\begin{algorithm}[H]
\caption{Character action selection process}
\label{alg:act}
\begin{algorithmic}[1]
\Function{Act}{Character $c$}
\State $score$ $\gets $ List of scores, defaulted to -1
\For{$i \gets 1$ to $|c.A_c|$}
\If{$\textit{Distance}(c.A_c[i], c.w_p) = 0$}
\label{alg:precon}
\State $score[i] \gets \text{Tension}_{\text{gs}}(c) - \text{Predict}(c, c.A_c[i])$
\label{alg:score}
\EndIf
\EndFor
\If{$max(score) >= 0$}
\Return $c.A_c[max(score)]$
\Else{}
\Return $null$
\EndIf
\EndFunction
\end{algorithmic}
\end{algorithm}

\begin{algorithm}[H]
\caption{Predict action change in subj. global tension}
\label{alg:prediction}
\begin{algorithmic}[1]
\Function{Predict}{Character $c,$ Action $a$}
\State $w_{pred} \gets $ Apply($a.w_\epsilon, c.w_p$) 
\State $tension \gets 0$
\For{\textbf{each} $w_v \in c.W_v$}
\State $tension \gets tension + \text{Distance}_s(c, w_v, w_{pred})$
\EndFor 

\State \textbf{return} $tension$
\EndFunction
\end{algorithmic}
\end{algorithm}

\begin{algorithm}[H]
\caption{Apply Postcondition to World}
\label{alg:apply}
\begin{algorithmic}[1]
\Function{Apply}{Effect $w_\epsilon,$ World $w$}
\State $w_n \gets w$
\For{$i \gets 1$ to $|w_n.V|$}
\If{$\neg w_\epsilon.V[i] = \bot$}
\State $w_n.V[i] \gets w_\epsilon.V[i]$
\EndIf
\EndFor 
\State \textbf{return} $w_n$
\EndFunction
\end{algorithmic}
\end{algorithm}

\subsection{Functionality}
\label{sec:functionality}

Considering finally the most foundational functioning of the system, we want a situation in which each character searches through their actions and greedily selects the action which most reduces their goal tension. This is a simple, generally deterministic approach to the system, which is likely to be unsuited to any significant EN work but nonetheless provides full functionality. \newchange{In a full EN work, authors would likely wish to include other features into action selection, such as memory, character relations, etc. We wish to avoid, however, making any assumptions about which features might be built upon the foundational model provided here, although adding such features can easily be implemented as part of the scoring function shown in Algorithm~\ref{alg:score}.} \change{Furthermore, this functionality is not used within the analysis we provide in this paper, however as we will discuss in Section~\ref{sec:conclusion} the functionality is useful in generating potential traces of the system that may be analyzed using techniques described in this paper.} In this functioning, the simulation occurs in a number of \textit{steps}, in which we iterate through all characters in the narrative and get them to select a valid (ie. preconditions met) action that best reduces the character's subjective goal tension. Each valid selected action is applied both to the actual world and each character's perceived actual world. The pseudocode for the step functionality is shown in Algorithm~\ref{alg:turn}, the action selection pseudocode shown in Algorithm~\ref{alg:act}, and the apply functionality shown in Algorithm~\ref{alg:apply}.

During each step, each character iterates through all their valid actions and \textit{predicts} what they believe the actual world will be like after that action. The ``score'' of an action is a measure of the change in goal tension between the predicted and current world. The prediction pseudocode is shown in Algorithm~\ref{alg:prediction} and essentially performs the goal tension calculation but on the predicted world. Note that the precondition checking for algorithms is determined subjectively by checking that the distance between the action and the perceived actual world is zero. This allows characters to select actions which are invalid in the actual world due to incorrect belief, hence we need to check the actual distance when taking the action, failing the action if necessary.  Characters may likewise select an action which is not actually ideal due to this same false belief, but characters will learn the actual truth values if they are influenced by an action, meaning that over time characters will progressively learn more and more of the true state of the actual world.

\section{Tension Space and Movement Analysis}
\label{sec:analysis}

In this section we define our analysis approach for the EN system and evaluate its ability to discover the presence of design decisions in the EN work \textit{Subject and Subjectivity}~\cite{subjectandsubjectivity} (\sas{}). All the subsequent work both this section and Section~\ref{sec:sketch} were performed using an authorship tool that fully implements the base system presented in Section~\ref{sec:model}, as well as the analysis and sketch tools defined in the remainder of this paper. The authorship tool was implemented in C\# in the Unity game engine. It is implemented as an extension to the Unity editor, allowing its use during the design phase of an EN work. %The tool allows for direct editing of the content, the analysis presented in this section, the sketch-based content design presented in Section~\ref{sec:sketch}, features to check for errors in content, and the option to playtest the narrative model using the base functionality described in Section~\ref{sec:functionality}.

\subsection{Tension Space and Movement}

The goals for our analysis is twofold. First, we want a measure and \newchange{means to visualize} \textit{tension}, the potential for conflict in the model. Second, we wish to analyze how the actions do or do not realize these different conflicts. For ease of reference we refer to the inverse of a conflict as a harmony.

\begin{figure}[ht]
\centering
\includegraphics[width=0.9\linewidth]{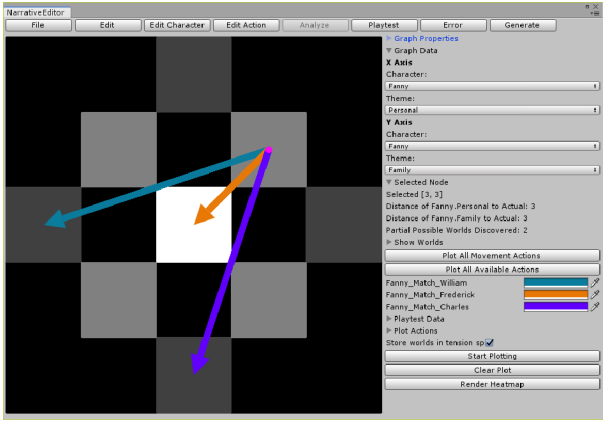}

\caption{The analysis tool, showing the tension space for the personal and family worldviews of Fanny from \sas{}, with the starting world highlighted with a pink dot and three available actions highlighted using three different coloured arrows.}
\label{fig:analysistool}
\end{figure}

Considering first the tension analysis, we previously mentioned that it is at the proposition where truth values differ in worldviews where tension is created, and further that for any given state of the actual world, the distance between a worldview and the actual world (or perceived actual world) gives a measure of the goal tension for that particular worldview. Thus, if we calculate the distance of this worldview to all possible states of the actual world, we get a set of possible goal tensions which can be true for that worldview, independent of what actions are actually possible. We call this the \textit{tension space} (TS) of the worldview, and is what we roughly treat as the expressive range of a given worldview. While useful, we more specifically want to evaluate the personal and interpersonal tensions which are present within the model. To do this, we define a two-dimensional tension space, which consists of one x-axis worldview and one y-axis worldview. This tension space is built by iterating through all possible states of the actual world, and plotting the goal tensions of both worlds using a heatmap\footnote{In actuality we perform several optimizations to better accommodate larger worlds, since the number of possible worlds grows exponentially with size and range.}. Thus for any given state of the actual world, we are at some (x,y) \textit{position} in the tension space, where x is the goal distance of the x-axis worldview to the state of the actual world and y is the same but for the y-axis worldview. In the heatmap, positions that are brighter mean that more possible states of the actual world lie at this position. An example of such a tension space can be seen in the heatmap in Figure~\ref{fig:analysistool}. We can use this same analysis for \change{different} worldviews of the same character to get personal tension spaces, or worldviews of \change{different} characters to get interpersonal tension spaces.

\begin{figure}[h!]
\centering
\begin{subfigure}{0.5\linewidth}
\includegraphics[width=\linewidth]{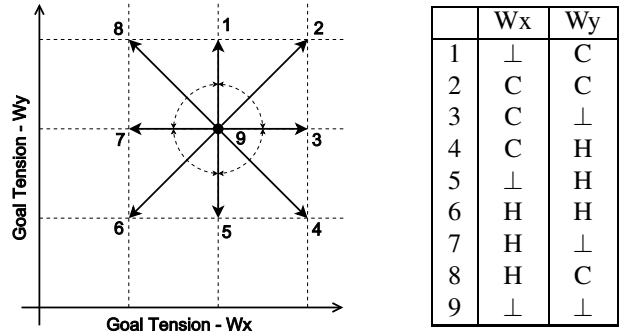}
\end{subfigure}
\hspace{0.1\linewidth}
\begin{subfigure}{0.3\linewidth}
\begin{tabular}{|c|c|c|}
\hline
& Wx & Wy\\
\hline
1 & $\bot$ & C\\
2 & C & C\\
3 & C & $\bot$ \\
4 & C & H\\
5 & $\bot$ & H\\
6 & H & H\\
7 & H & $\bot$ \\
8 & H & C\\
9 & $\bot$ & $\bot$ \\
\hline
\end{tabular}
\end{subfigure}
\caption{Diagram showing the eight \change{movements} that two worlds, \textit{Wx} and \textit{Wy} can move in, and a chart showing the results for each world (C = Conflict, H = Harmony, $\bot$ = \change{``don't care''}). \change{In the remainder of this paper we will refer to movements using the legend on the right (e.g. a 2 movement relates to a conflict for both worlds.)}}
\label{fig:movement}
\end{figure}

At any position in the tension space, there are certain actions which will be available for one or more of the possible actual worlds at that position. Since the actions will change this state of the actual world to a new one, an action can be thought of as a \textit{movement} in the tension space, from one position to the other. Given that the overarching goal of each character is to have the distance of all worldviews to the actual world be zero, we can say that any movement towards the zero position on an axis in the plot is harmonious for that axis' worldview, and any movement away is conflicting. An action which causes no movement in an axis, indicates that for that worldview, all affected truth values must be \change{``don't care''}. From this, we can classify all possible movements in the tension space according to their affect on the x and y-axis worlds. This classification is shown in Figure~\ref{fig:movement}, and in the remainder of the paper when we refer to eg. a 6 movement we a refering to the labelling in this figure. Notably, when considering the foundational functionality, only movements 6, 7 and 8 make sense as possible action choices for the x-axis worldview, whereas movements 4, 5 and 6 are similar for the y-axis worldview.

Note that the tension space does not represent the actual space available to the interactor. Only a subset of the tension space may be reached through actions, and often the accessible portion of the tension space is radically smaller. From an authoring perspective, the intent of the tension space is to assist in the design of new actions, showing which areas are possibly accessible, and allowing the author to visualize the impact of an action on a given set of worldviews in terms of its movement. Figure~\ref{fig:analysistool} shows a complete example of a tension space with available movements from \sas{}~with the starting world selected (the magenta dot) and the three arrows showing the possible movements occurring from the three possible actions. %For a given narrative, the actual world used in $N$ is always marked using a magenta dot, and any point can be clicked by the author to see all possible actual worlds at that point, and even render which actions are available at that world.

Tension space also provides useful feedback for authors when trying to predict the behaviour of the system, in that certain tension spaces better afford certain movements, and the shape of this space thus defines a high-level relation between worlds. As an example, a tension space that is roughly diagonal in the direction of the 4 and 8 movements, implies that most actions in this space will \change{create} either a 4 or 8 movement, ie. most actions will harmonize for one worldview, and conflict for the other. This means that most actions will cause conflict for at least one world, and thus we call this shape a \textit{strong} tension space, as the potential for conflicts between worldviews is high. \change{Note that this is not ensured, and in fact all actions in a strong tension space may actually occur in the 2 or 6 direction. However, without making any assumptions about the actions, we are merely interested in whether there is a much higher potential for conflicting actions.} Conversely, we call a 2 to 6 shaped tension space a \textit{weak} tension space, since an action which is harmonious for one world is generally harmonious for the other. Examples of both such spaces are shown in Figure~\ref{fig:tensionspace} and explained further in Section~\ref{sec:sasanalysis}. Also important is that the bounds of the tension space define the limits of cooperation or conflict between worldviews. Only in cases where the 0,0 position is accessible and reachable through actions can the two worldviews simultaneously achieve their overarching goal. Otherwise there is always a diagonal lower bound line along which only 4 and 8 movements are possible, and thus conflict for at least one worldview is inevitable. Likewise the diagonal upper-bound indicates the same but instead for the maximum tension. The further away the lower-bound is from the bottom-left, the more the satisfaction of one worldview occurs at the cost of the other, and is a prominent feature of strong tension spaces.

The tool likewise allows the ability to visualize a trace of the system, by drawing the movements taken by each action, giving a sequence of movements that goes from the starting position to some ending position. Though not explored in this paper, future work aims to investigate if the shapes of these movements relates to particular structures of stories.% This knowledge would allow the detection of higher level structures within tension space during authorship, but further holds application to drama management~\cite{robertsDramaManagementSurvey2008} and story curation~\cite{jryanphd}, both of which require a knowledge of story structure to work.
%The tool allows both the ability to see all actions from a position in the tension space, or to visualize a trace from a playthrough of the EN using base functionality. These playthroughs can be done either manually selecting actions or simply running the base functionality for all characters up to a certain length.

\subsection{Subject and Subjectivity Analysis}
\label{sec:sasanalysis}

\begin{figure*}
\centering

\begin{subfigure}{0.3\linewidth}
\begin{tabular}{| m{2em} | c c c |}
\hline
 & Pers. & Fam. & Soc.\\
\hline
 Pers.
 & \includegraphics[scale=5]{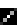}
 &  \includegraphics[scale=5]{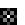}
 & \includegraphics[scale=5]{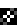} \\

 Fam.
 & \includegraphics[scale=5]{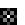} 
 &  \includegraphics[scale=5]{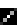}
 & \includegraphics[scale=5]{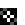} \\
 
  Soc. 
 & \includegraphics[scale=5]{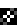} 
 &  \includegraphics[scale=5]{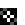}
 & \includegraphics[scale=5]{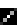} \\
\hline

\end{tabular}
\caption{Fanny}
\label{fig:tsfanny}
\end{subfigure}
\begin{subfigure}{0.3\linewidth}
\begin{tabular}{| m{2em} | c c c |}
\hline
 & Pers. & Fam. & Soc.\\
\hline
 Pers.
 & \includegraphics[scale=5]{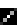}
 &  \includegraphics[scale=5]{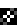}
 & \includegraphics[scale=5]{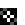} \\

 Fam.
 & \includegraphics[scale=5]{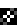} 
 &  \includegraphics[scale=5]{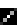}
 & \includegraphics[scale=5]{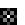} \\
 
  Soc. 
 & \includegraphics[scale=5]{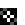} 
 &  \includegraphics[scale=5]{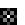}
 & \includegraphics[scale=5]{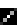} \\
\hline

\end{tabular}
\caption{Jane}
\label{fig:tsjane}
\end{subfigure}
\begin{subfigure}{0.3\linewidth}
\begin{tabular}{| m{2em} | c c c |}
\hline
 & Pers. & Fam. & Soc.\\
\hline
 Pers.
 & \includegraphics[scale=5]{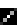}
 &  \includegraphics[scale=5]{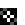}
 & \includegraphics[scale=5]{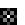} \\

 Fam.
 & \includegraphics[scale=5]{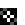} 
 &  \includegraphics[scale=5]{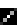}
 & \includegraphics[scale=5]{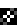} \\
 
  Soc. 
 & \includegraphics[scale=5]{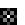} 
 &  \includegraphics[scale=5]{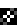}
 & \includegraphics[scale=5]{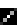} \\
\hline

\end{tabular}
\caption{Elizabeth}
\label{fig:tselizabeth}
\end{subfigure}

\caption{Tension space for each character's internal worldviews in the game \textit{Subject and Subjectivity}. Pers., Fam. and Soc. refer to the Personal, Family and Society themes respectively}
\label{fig:tensionspace}
\end{figure*}

In considering the evaluation of tension space and movement analysis within an actual context, we here evaluate a work created in the system, \textit{Subject and Subjectivity}. The work was created before tension space analysis or even the formalization of the model, but nonetheless was designed in such a way to realize a set of strict design goals surrounding personal tension. Although we authored the game ourselves, it is nonetheless valuable to investigate if through this new form of analysis we can visually see these design goals through tension space shape and action movement.
%Here, we apply tension space and movement analysis to \textit{Subject and Subjectivity}, which is a work created in the system, and show how certain of its design goals can be visually represented using this analysis. As previously mentioned, \sas{} is used here simply because it is a work made using the model but designed without any use of the authoring tool or tension space analysis, but was nonetheless designed with a very specific structure of internal character tension and is additionally small enough to be fully modelled.

\begin{figure}[htbp]
\begin{subfigure}{\linewidth}
\centering
\begin{tabular}{|r|c|c|c|c|}
\hline
Character & Theme & Wealth & Faith & Ambition \\
\hline
\multirow{3}{*}{Fanny} & Personal & T & T & F\\
& Family & F & T & T\\
& Society & T & F & F\\
\hline
\multirow{3}{*}{Jane} & Personal & F & F & T\\
& Family & T & F & T\\
& Society & T & T & F\\
\hline
\multirow{3}{*}{Elizabeth} & Personal & T & F & T\\
& Family & F & T & F\\
& Society & F & T & T\\
\hline
\end{tabular}
\caption{The character's internal worldviews.}
\label{tab:saschars}
\end{subfigure}

\begin{subfigure}{\linewidth}
\centering
\begin{tabular}{|r|c|c|c|}
\hline
Bachelor & Wealth & Faith & Ambition \\
\hline
William (Fanny) & T & T & F\\
Frederick (Jane) & T & F & T\\
Charles (Elizabeth) & F & T & T\\
\hline
\end{tabular}
\caption{The bachelor's effects on the actual world. The best match for each bachelor is shown in brackets.}
\label{tab:sasbach}
\end{subfigure}

\caption{The character and bachelor models used in \sas.}
\label{tab:sasmodels}
\end{figure}

\textit{Subject and Subjectivity} (\sas) is a conversational puzzle game designed using the possible worlds model. Designed as a loosely Jane Austen inspired matchmaking game, the interactor is tasked with matching each of their three friends with a bachelor, balancing the friend's personal interests and family and societal obligations. Systemically, the narrative model has a true/false range (0, 1), three characters: [Fanny, Jane\newchange{,} Elizabeth], three themes: [personal, family, society], and four propositions per character (for 12 in total): are they [matched] and is the character paired with someone who is [wealthy, ambitious\change{,} and religious]. There are three actions for each character to match that character with a bachelor, which sets the matched value to true and the remaining three propositions are set to the values of that bachelor (ie. are they weatlhy, ambitious or religious). There is further one action per character to unmatch with a bachelor and return to the starting state. Thus matching a character with a bachelor means that their particular set of propositions in the actual worlds will now match that of the bachelor. The puzzle of the game is that only one bachelor per character is ideal for reducing the overall goal tension, and the player must discover which bachelor this is while only having limited information about each character worldviews through conversation.

%\sas{} is interesting to analyze mainly because, given that it was intended as a puzzle, it had clearly defined design decisions. First, each character needed to have three different worldviews, with two being more similar and two being different, and that the similar/different matchings are different for each character. Second, each bachelor must be different, and there should only be one ideal bachelor per character, one bachelor who \textit{appears} ideal in one worldview (a ``red herring'' character) but is overall not ideal, and one bachelor who never appears ideal, but is ``equally'' ideal to the second bachelor. Third, no characters should have a matching worldview for any theme and they should aim to have worldviews that are as different as possible. Fourth, the ideal bachelor is always perfectly ideal in one worldview, but this world is different for each character. Notably, no character cares about other character's bachelors, thus conflicts are only personal and not interpersonal. The final values were selected using a brute force check of all possible game models, and a summary of the resulting model of character and bachelor values is shown in Figure~\ref{tab:sasmodels}. 
\sas{} is interesting to analyze mainly because, given that it was intended as a puzzle, it had clearly defined design goals, three for worldviews and four for bachelors. The three \newchange{worldview design} goals were:
\begin{enumerate}
\item Each character has two worldviews with different truth values, the choice of the two worldviews is different for each character.
\item Each character has two worldviews with similar truth values, the choice of the two worldviews is different for each character.
\item Generally, ideals should be as different as possible for each character to make each character feel different
\end{enumerate}

The \newchange{bachelor (action) design goals} were:
\begin{enumerate}
\item One bachelor is ``ideal'' and reduces overall goal tension more than the other two bachelors (practically, this is -5, vs. -3 for the other two).
\item The ideal bachelor reduces goal tension to zero for one worldview.
\item One non-ideal bachelor also reduces goal tension to zero for a different worldview, as a ``red herring''.
\item The remaining bachelor is an equal reduction in goal tension to the second bachelor.
\end{enumerate}
Note that the arrangement of ideal/red herring/other bachelor is different for each character. No character cares about other character's bachelors, thus conflicts are only personal and not interpersonal. The final values were selected using a brute force check of all possible game models, and a summary of the resulting model of character and bachelor values is shown in Figure~\ref{tab:sasmodels}. 

In this experiment, we'll try to discover the worldview restrictions using tension space analysis, and the bachelor restrictions through movement analysis. Considering first the worldviews, Figure~\ref{fig:tensionspace} shows all possible internal tension spaces \change{for all} three characters. Since characters do not use \change{``don't care''} values for their personal bachelors, all possible movement is diagonal giving the tension spaces a ``checkerboard'' look. \change{This is because without ``don't cares'', it is impossible for the 1, 3, 5, 7 or 9 movements to occur, so the only remaining movements possible are 2, 4, 6 or 8 regardless of the structure of the action.} Note also that the space is diagonally symmetric, meaning that the shape TS for two worlds is the same regardless of which world is the x or y axis (this is not always the case). Further, if the same world is used for both axes, the result is always the weakest possible TS since a world cannot conflict and harmonize with itself at the same time so only 4,8 movements are possible. 

Considering the first worldview design goal, we expect to find a strong tension space for each character for a different set of two worldviews. This is notably the case in the family/society TS for Fanny, the personal/society TS for Jane, and the personal/family TS for Elizabeth. The second goal is similarly noted in the weak tension spaces of personal/society of Fanny, personal/family for Jane family/society for Elizabeth. The third design goal, interestingly, cannot be evaluated using the base model, as this would be revealed through interpersonal tensions, but since each character has \change{``don't care''} for each other's bachelors, no tension spaces exist.

\begin{figure}[htbp]
\centering
\begin{tabular}{| r | c c c |}
\hline
 & Pers./Fam. & Fam./Soc. & Soc./Pers.\\
\hline
 Fanny
 & \includegraphics[scale=0.1]{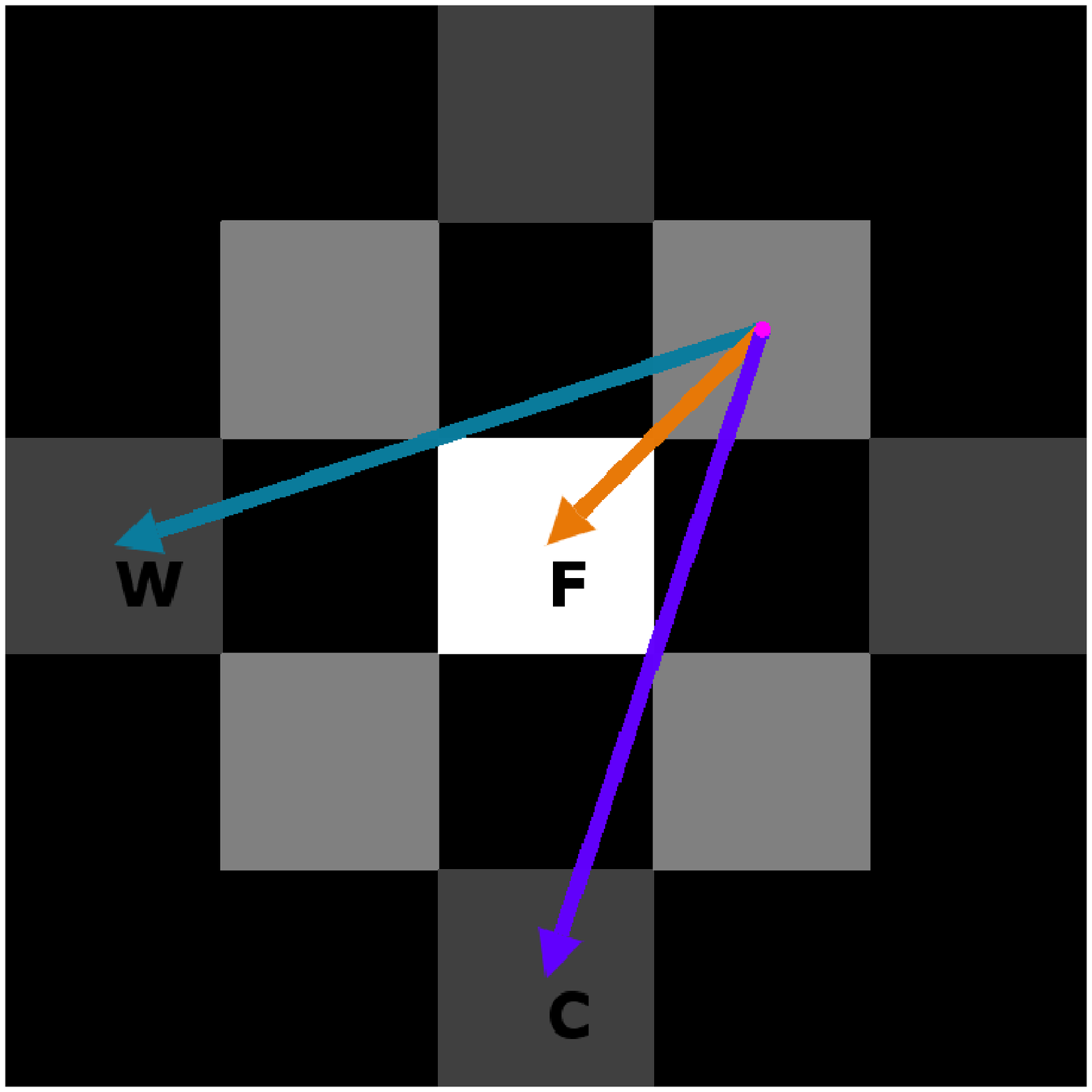}
 & \includegraphics[scale=0.1]{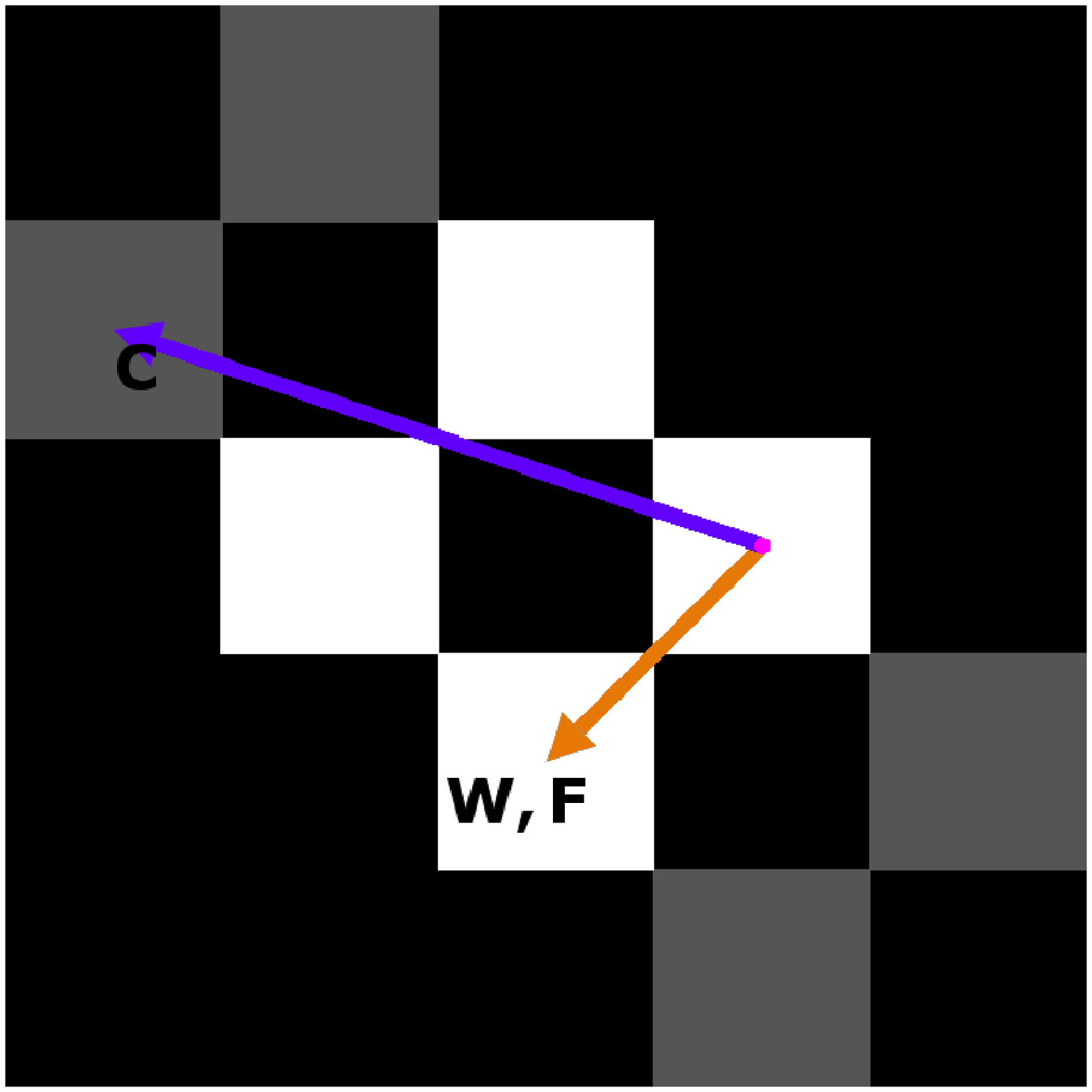}
 & \includegraphics[scale=0.1]{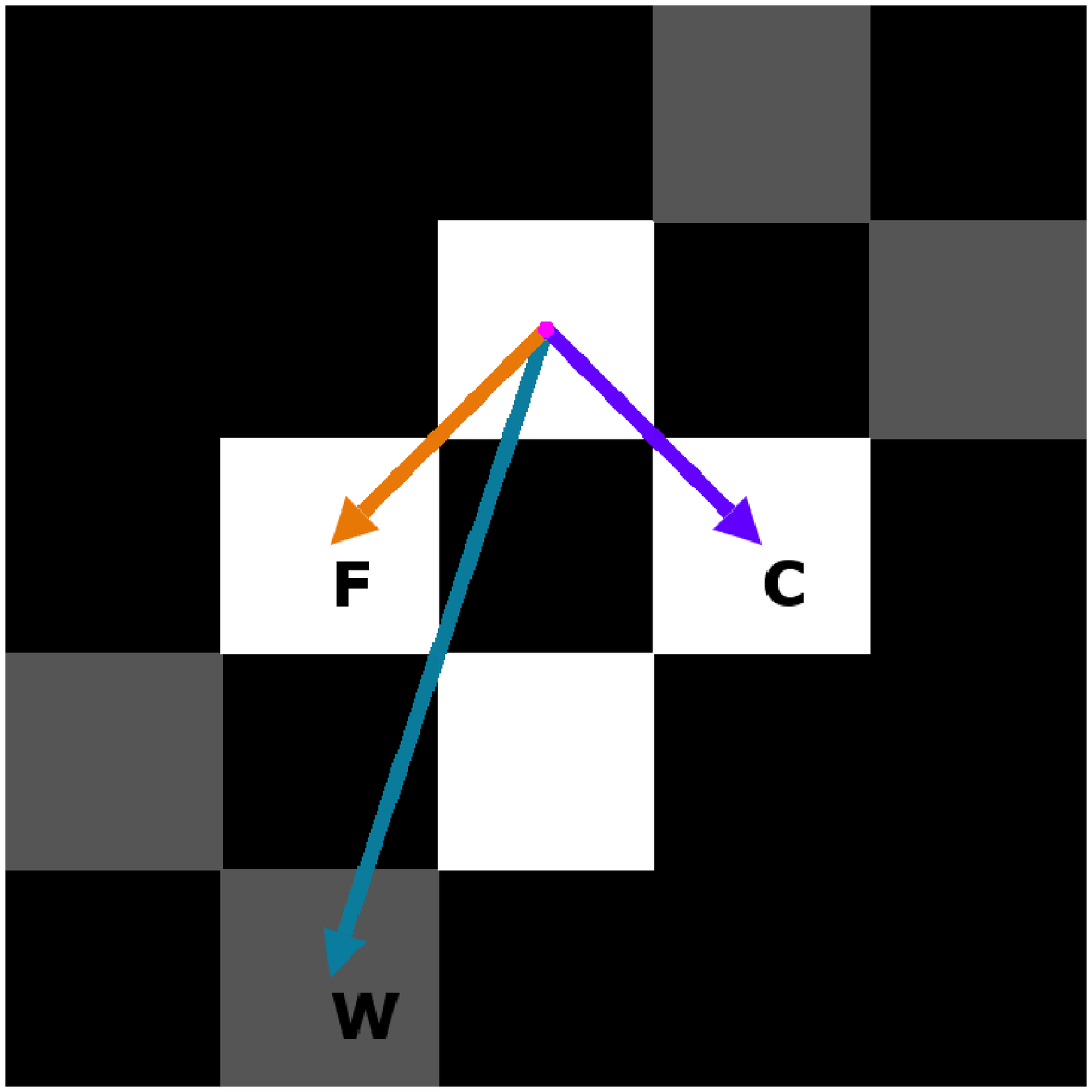} \\

 Jane
 & \includegraphics[scale=0.1]{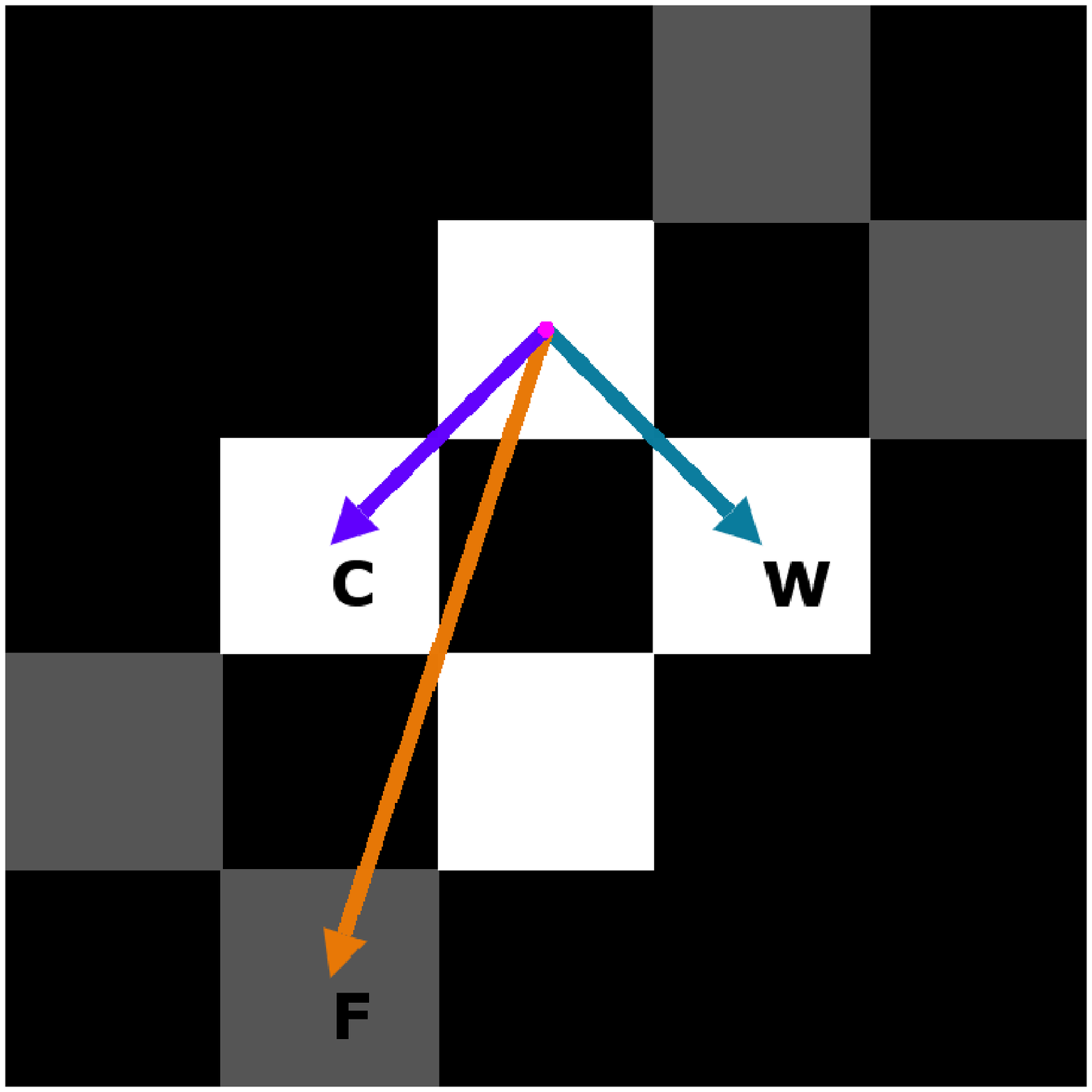}
 & \includegraphics[scale=0.1]{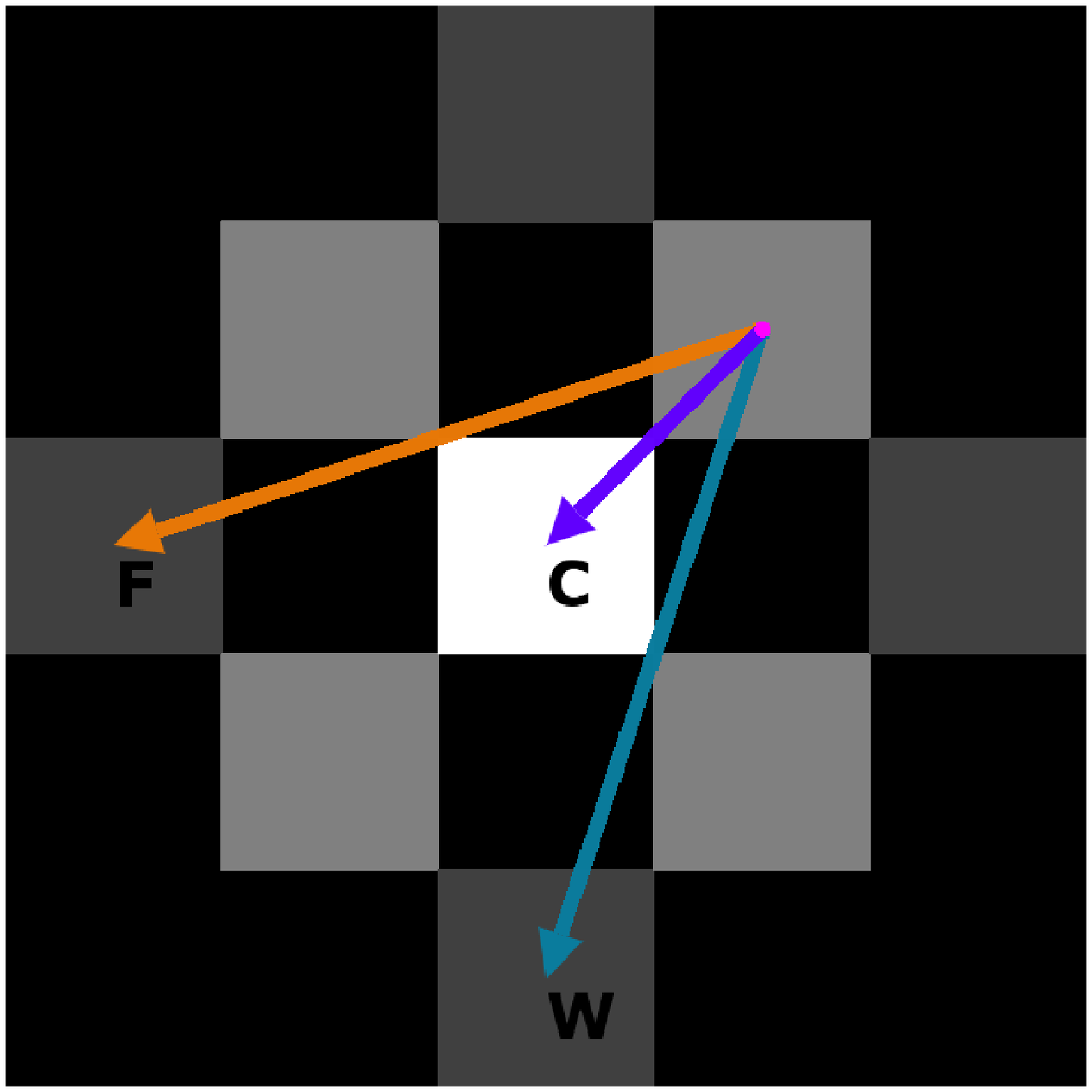}
 & \includegraphics[scale=0.1]{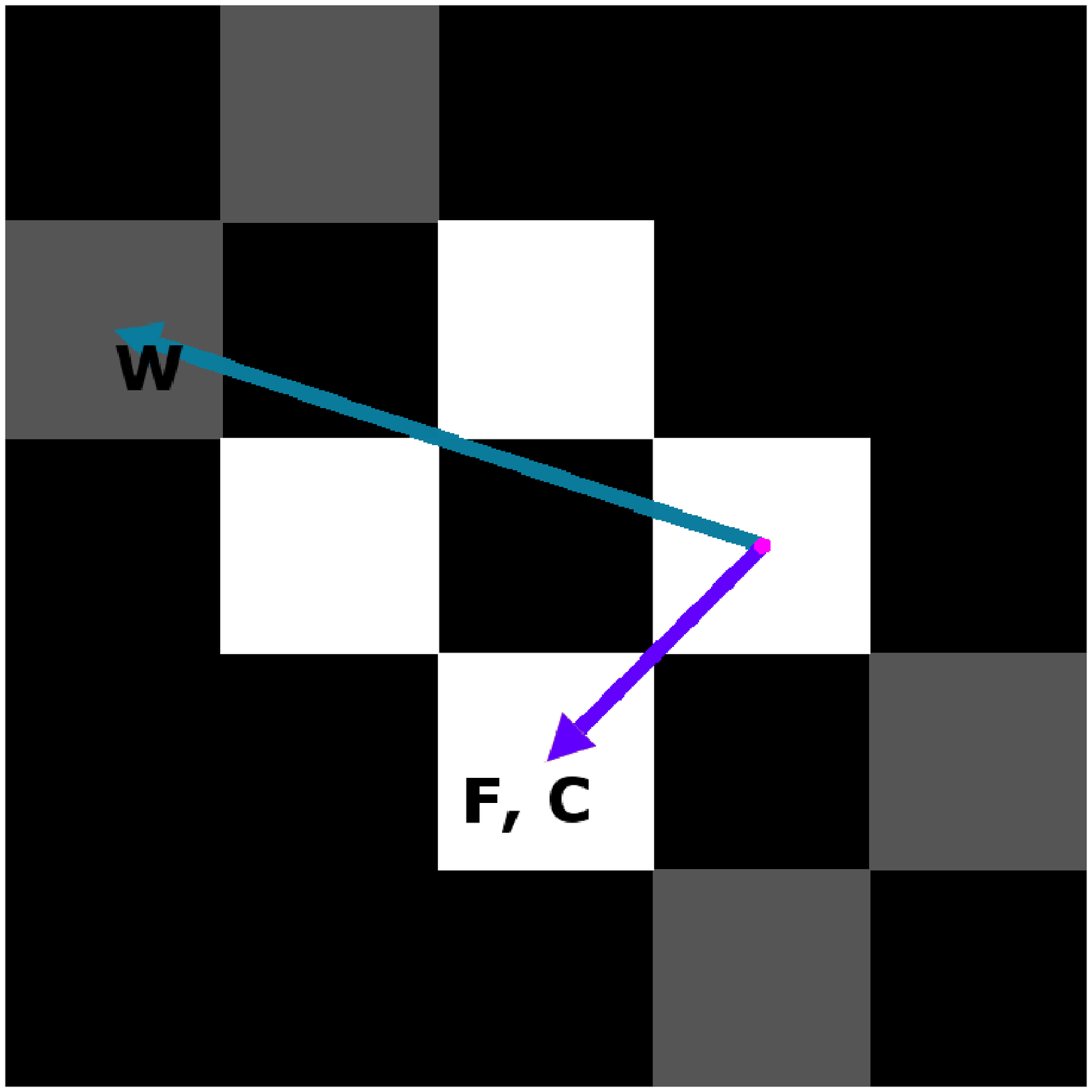} \\
 
 Elizabeth 
 & \includegraphics[scale=0.1]{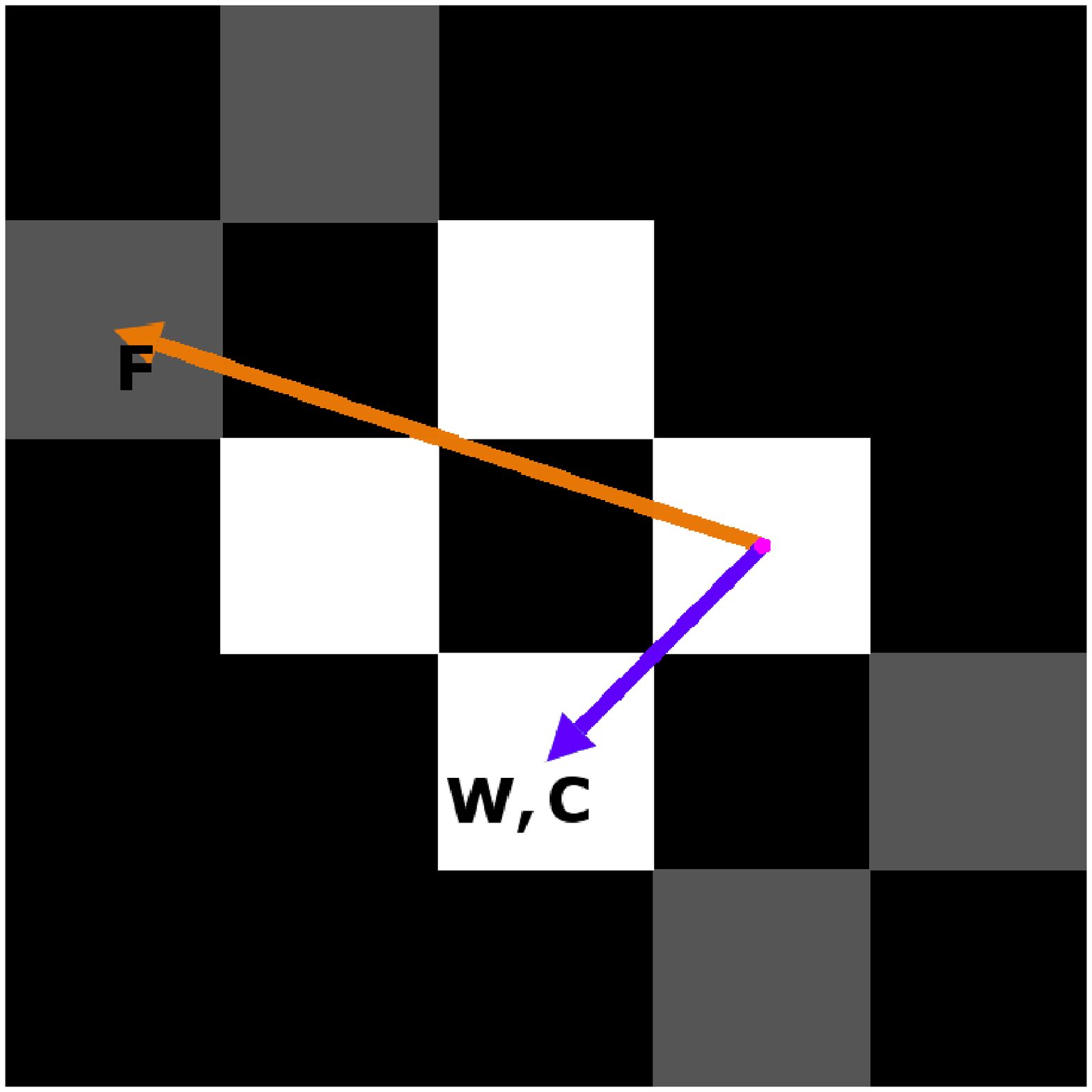}
 & \includegraphics[scale=0.1]{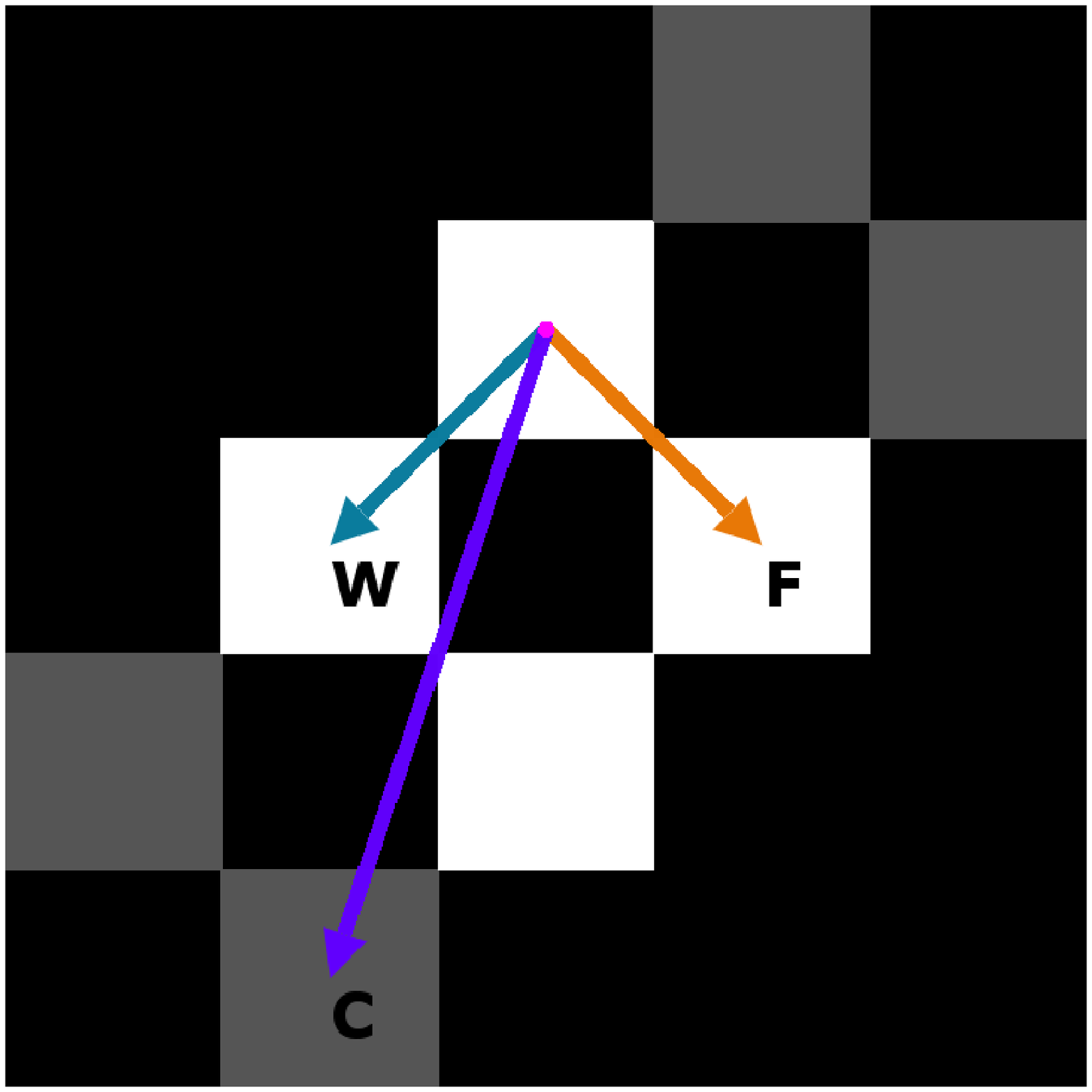}
 & \includegraphics[scale=0.1]{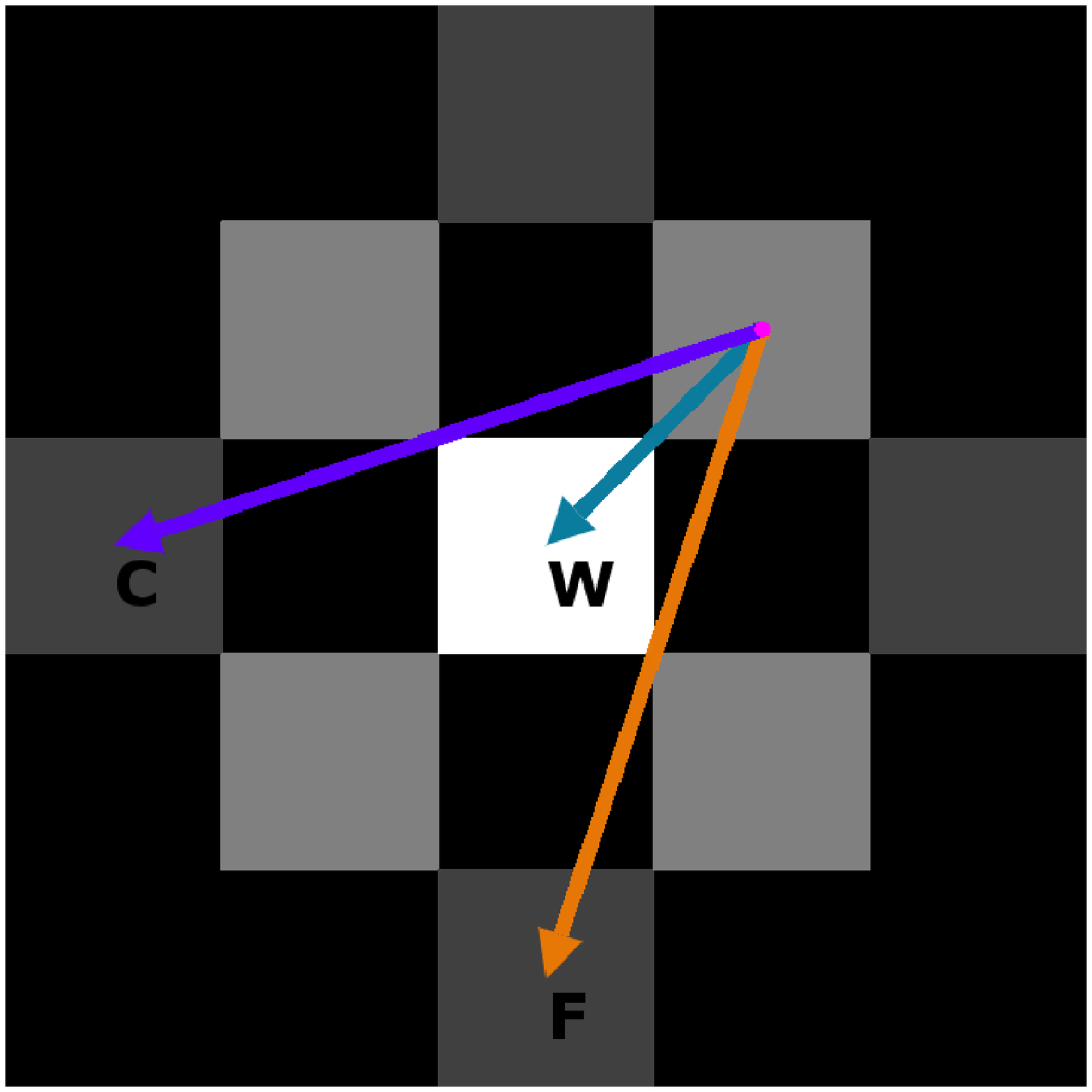} \\
\hline

\end{tabular}
\caption{Bachelor movements for the main three tension spaces.}
\label{fig:sasmovements}
\end{figure}

To evaluate the bachelor movement goals, we first look at the three main sets of worldviews: personal/family, family/society and society/personal, and plot the movements from the starting state according to each of the three bachelors. The results are shown in Figure~\ref{fig:sasmovements}, with each arrow being labelled and coloured according to action (both bachelors are listed in cases of overlapping movements). Interestingly, as with the tension spaces all three characters are essentially permutations on the same structure, which makes sense as the design goals essentially revolve around creating one type of character. Considering the first action design goal, the ideal bachelor in each case is always a viable 6 movement, and in the weak tension spaces is always the most ideal action. For the second design goal, we can see in the weak tension space that the movement always ends up at zero in one axis, and same with the ``diamond'' shaped axis. For the third design goal, the ``red herring'' character is always an 8 movement in the strong conflict space, and a 4 movement in the weak conflict space. There is only one tension space, the ``diamond'' shaped one, in which the red herring character has a doubly harmonious 6 movement, whereas the ideal bachelor has a similarly harmonious movement in equal magnitude. Likewise, we can see the axis along which the red herring character reaches zero in diamond and strong tension spaces. Lastly, the remaining bachelor \change{always has} a 6 movement, but always ever with magnitude 1, thus never having the dramatic goal tension reduction of the other two bachelors. Though not immediately obvious, calculating the precise overall goal reduction of each bachelor to check design goals one and four is also possible, by adding together the vectors of each movement, which gives a [-5, -5] movement for the ideal bachelor, and a [-3, -3] movement for the remaining two.

The analysis here was a positive result, being able to find patterns even for the very specific design goals outlined for the puzzle. In general, most analysis will likely be higher level, looking to detect general tensions which exist or the overall impact of actions, but as we've seen here with a strong understanding of the properties of tension space and movement, deep design patterns start to emerge. Further, by performing such an analysis we can open up directions for future work, such as trying to detect certain higher-level patterns, or even analyzing a given model to say which design goals it embodies.

\section{Tension Graph Sketching}
\label{sec:sketch}

\begin{figure*}
\centering
\begin{subfigure}{0.3\linewidth}
\centering
\includegraphics[width=\linewidth]{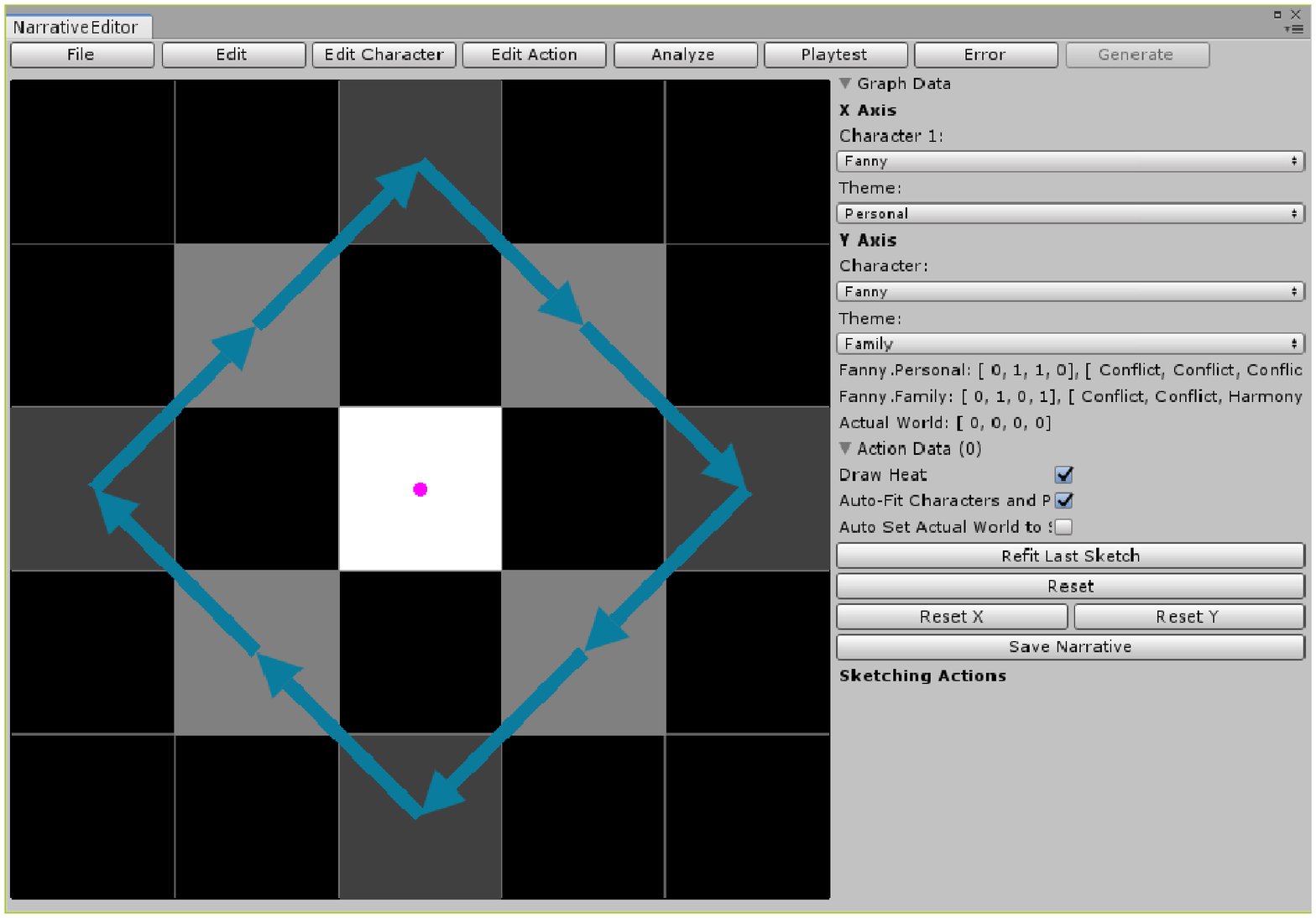}
\caption{Sketch to define the tension space between the personal/family worldviews.}
\label{fig:sketch1}
\end{subfigure}
\begin{subfigure}{0.3\linewidth}
\centering
\includegraphics[width=\linewidth]{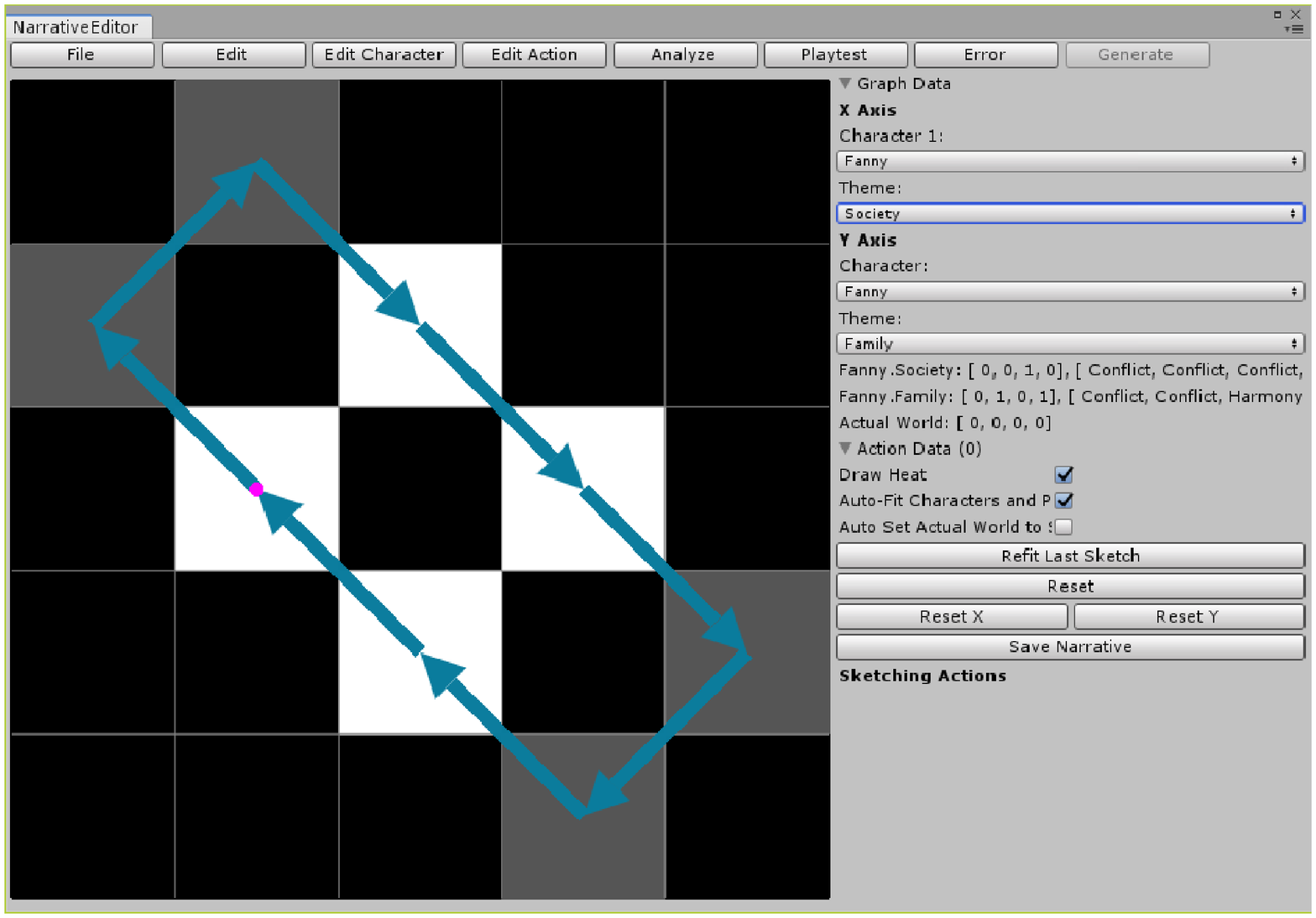}
\caption{Sketch to define the tension space between the society/family worldviews.}
\label{fig:sketch2}
\end{subfigure}
\begin{subfigure}{0.3\linewidth}
\centering
\includegraphics[width=\linewidth]{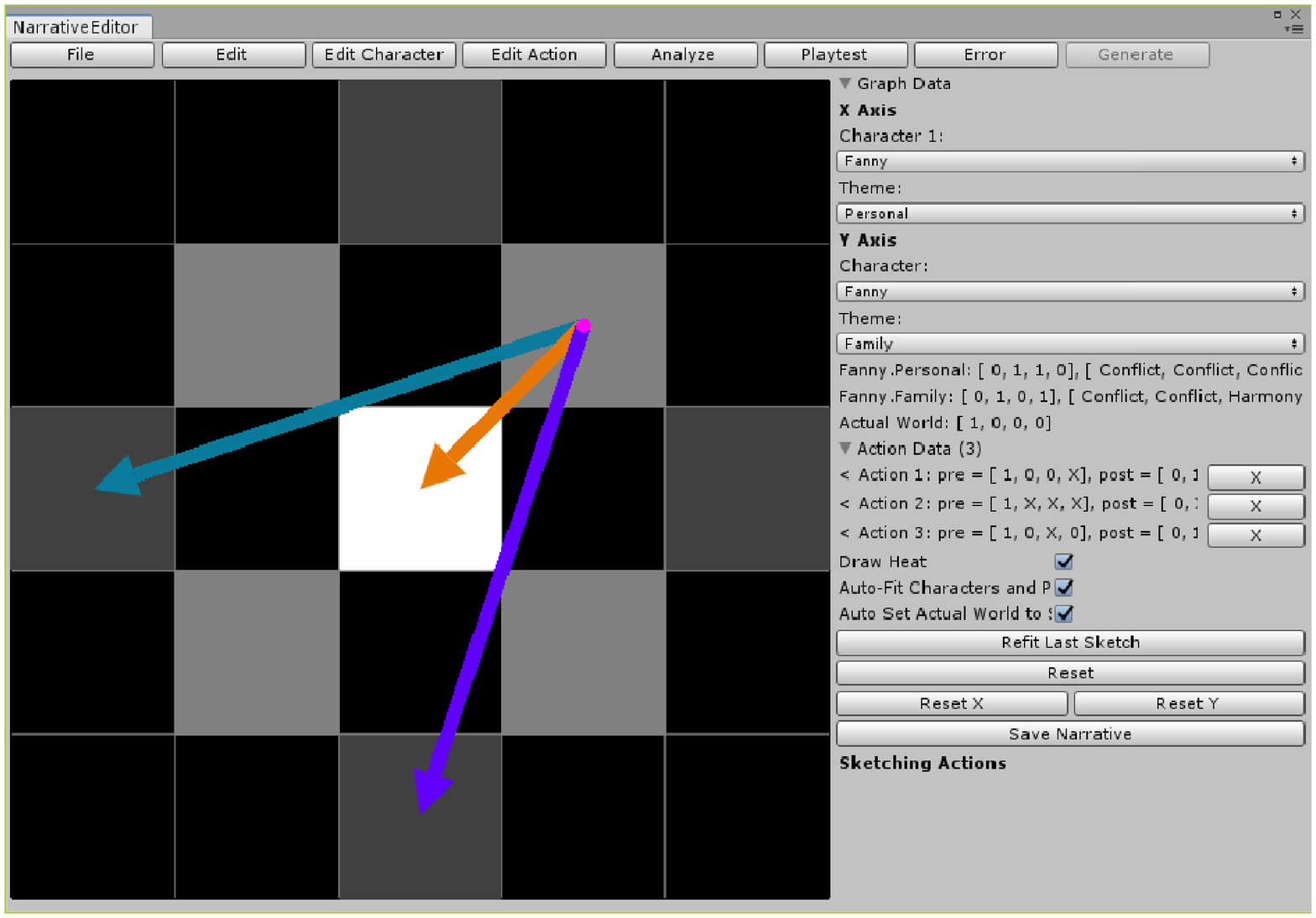}
\caption{Three sketches, for each action, in the personal/family tension space.}
\label{fig:sketch3}
\end{subfigure}

\caption{Screenshots of the three major steps in sketching the character Fanny from \sas{}, using five sketches total. \change{Figures~\ref{fig:sketch1} and \ref{fig:sketch2} are both sketches of the \textit{worldviews}, while Figure~\ref{fig:sketch3} is an \textit{action} sketch with each of the three sketches coloured for comparison purposes to the actions we are trying to replicate (which may be seen in Figure~\ref{fig:sasmovements}).} Colors here, as with analysis, do not influence generation but can be used to differentiate different sketches.}
\label{fig:sketchprocess}
\end{figure*}

Tension space analysis suggests one potential direction for authorship, which focuses first on defining the tension space through character worldviews, and then defining the intended movements through this space with actions. Further, if we ignore any particular classification of propositions or actions in advance, it is possible to author an emergent narrative specifically through the creation of tension spaces and movements, and then fitting a context onto this structure. This is an inverted approach to traditional EN authorship, which often starts with a concept or genre of narrative that is de-constructed into actions and characters.

In this section, we experiment with the possibility of using a sketch-based approach to authoring, where an author is able to ``draw'' tension spaces and movements that result in the complete foundational model of narrative presented in Section~\ref{sec:model}. This approach focuses on conflict design, constructing the desired ways in which characters will harmonize or conflict, and the way they are able to move about the space. Our goal in this is to demonstrate the practical potential
%While the precise impact of such on authoring tool is unknown, in this paper we aim mainly to prove the possibility 
of such an approach, using a sketch-based metaphor for designing tension spaces and actions. As an evaluation, we return to our previous section's analysis of \sas{}, although this time we aim to see if we can recreate the character structure from above, and how many of the original design goals are met. At this stage, the sketch tool works only with true/false ranges, however it is fairly straightforward future work to extend the approach to larger ranges.

%Experimentally, we want to ensure that such an approach does allow the author to embed structural design goals into their authorship, and as such following the description of the sketch-based approach we attempt to reconstruct a character structure and actions from \sas{}, comparing the results with those obtained in Section~\ref{sec:sasanalysis}. Notably, at this stage we focus specifically upon a true/false range, though it is fairly straightforward to extend the same functionality to higher ranges.

\subsection{The Sketch Interface}

The sketch interface works upon two basic principles. First, we can treat any sketch in a grid layout as a sequence of movements, with each movement being one of the nine movements shown in Figure~\ref{fig:movement}. \change{We define a \textit{unit movement} as a movement in one of the directions from \ref{fig:movement} (excluding movement 9 as it involves no movement), which occurs when only one value in the actual world is changed. An action which changes multiple world values can therefore be deconstructed into a set of unit movements, one for each truth value affected. For example in \sas{}, matching with a bachelor affects four truth values, which means this action can be decomposed into four discrete unit movements.} 
%Furthermore, changing one truth value in the actual world will result in a \textit{unit movement} in one direction (excluding movement 9). 
Thus, even complex actions such as the bachelor matching actions in \sas{} can be decomposed into a set of unit movements, one for each truth value, that result in the larger movement of the action. Second, each movement defines a relation between the worldviews, and so long as the truth values respect this relation, we can assign them to the worldviews. For example, an 8 movement is caused by a harmony in the x-axis world and a conflict in the y-axis world. This implies that the x-axis and y-axis world must have different values, and further that there exists an action which modifies this truth value in such a way that it moves closer towards the x-axis world's truth value and further from the y-axis' value. Considering a true/false range, we have two equivalent options, either x is true, y is false and there is an action moving a truth value from false to true, or vice versa where x is false, y is true and there is an action moving a truth value from true to false. Both of these possibilities are structurally equivalent, thus allowing a variety of ways to realize a given sketch.

From this perspective, given any sketch drawn within a grid (we use the same heatmap layout as the analysis tool), we can \textit{fit} the sketch to either worldview truth values or actions. \change{We thus define \textit{fitting} as taking a sketch of either worldviews or actions, and attempting to create or modify a worldview/action to ``fit'' the unit movement decomposition of that sketch.} The basic fitting process consists of (1) having the user make a sketch in the form of a series of large movements, which are (2) decomposed into a set of unit movements that are (3) fit either to worldviews or to an action. To fit worldviews \change{we simply iterate} through all unit movements, look at the relation between worlds implied by the movement, and then randomly fit a set of possible values which obey said relation. For actions, we treat each large movement as an action, and treat each of the decomposed unit movements as one truth value modified by the action.
%, using the worldview relations to fit possible values via an A* algorithm search. Pseudocode for both worldview and action fitting procedures are shown in Algorithm~\ref{alg:fitTS}

For our sketch-tool we treat the fitting of worldviews and actions separately, meaning an author first has to sketch the tension space, and can then sketch to create actions. There are two reasons for this split. First, when fitting worldviews and actions at the same time, the tension space will closely resemble the actions created; however in many cases, such as \sas{}, the actual shape of the tension space is very different than that of the actions. Second, and more importantly, the starting position of an action is important. Essentially, an action starts at some state of the actual world and causes changes to it. By having a fully defined tension space, we can find a possible actual world at the start of the action, and use that to set the preconditions for the action, and this allows us to treat an action sketch as a desirable trace of the system. This is especially useful in defining actions starting from the intended starting state of the actual world in the narrative. The sketch process for both fitting procedures is always the same, with the author drawing lines that link together nodes in the grid with an edge, which gets stored as a movement and a unit decomposition. The tool then attempts to fit the sketch as appropriate.

The worldview fitting procedure supports partially defined worldviews. Essentially, if some values are defined in one or both of the worlds, then the system will treat these as restrictions upon the possible movements for any defined truth values. For example, if an x-axis worldview truth value is defined as true, then the y-axis worldview must be true if having a matching relation, or false for differing, and further that any movements that require the x-axis to be \change{``don't care''} (1,5 and 9) are impossible for this truth value. This is especially useful if we have a defined worldview, and want to fit a second worldview in relation to it, as we will do in Section~\ref{sec:chardesign}.

\change{\textit{Failures} occur when one or more of the unit movements made in the sketch cannot be fit to either a worldview or action, thus it is not possible to perfectly fit the sketch. During each stage, these failures are handled differently.} At the worldview fitting stage, the sketch tool attempts to fit as many of the unit movements of the sketch as possible, ignoring those that cannot be fit. For actions, since we treat the sketch as a trace for the system, we attempt to fit the movements sequentially, keeping track of the updated state of the actual world, and if one action fails we stop and do not fit any further movements (these failures are labelled in red for the author). 

\begin{algorithm}[H]
\caption{Fit Sketch As Tension Space}
\label{alg:fitTS}
\begin{algorithmic}[1]
\Function{FitTS}{Sketch $s$, World $w_x$, World $w_y$}
\For{\textbf{each} Edge $e \in s.E$ }
\State $a \gets $ Assigned proposition for each movement
\For{$i \gets 1$ to $|e.M|$}
\State $a[i] = $ Best fit of $e.M[i]$
\EndFor
%\For{$i \gets 1$ to $|w_n.V|$}
%\If{$\neg w_\epsilon.V[i] = \bot$}
%\State $w_n.V[i] \gets w_\epsilon.V[i]$
%\EndIf
\EndFor 
\If {all movements assigned}
\For{$i \gets 1$ to $|e.M|$}
\State Fit $w_x.V[a[i]]$ and $w_y.V[a[i]]$ with $e.M[1]$
\EndFor
\EndIf
\EndFunction
\end{algorithmic}
\end{algorithm}

Technically, a \textbf{sketch} $s$ is defined simply as $s = \langle E \rangle$, where $E$ is a set of \textit{edges}. An \textbf{edge} $e$ is a line segment connecting two nodes in the grid and is defined as $e = \langle n_{s}, n_{e}, M \rangle$ where $n_{s}$ is the \textit{start node}, $n_{e}$ is the \textit{end node}, and $M$ is a decomposition of the edge into a set of \textit{unit movements}. Decomposition is done by taking the slope between $n_{s}$ and $n_{e}$ and then selecting one possible set of movements which results in the change of distance indicated by the slope.

Given some sketch, $s$, the fitting procedure consists of fitting each movement of each edge to a proposition in the worldview, but the way this fitting works depends on the stage. \change{For the tension space sketching phase,} Algorithm~\ref{alg:fitTS} shows the steps in fitting the sketch to the tension space. The algorithm essentially proceeds through all edge movements and assigns them iteratively. If each edge movement can be mapped to one proposition, then the two worldview's proposition will be assigned a truth value based upon relation defined by the movement.

For the second stage, in which we fit each edge to an action, we first attempt to find a possible world state at the start of the sketch, and return a failure if none is found. From this starting world state, we attempt to find a sequence of actions to reach the subsequent end nodes of each edge using the A* search algorithm. Essentially, from our starting state, each of the propositions can be one possible unit movement, so we can treat this as our total possible set of movements for that action, and search from there, using the Manhattan distance between nodes as our heuristic score. If we reach the end node, then we take the list of movements we used to get to the node, and which propositions they correspond to, and then generate an action which has those movements as the precondition and their changes as the postcondition. We then apply this action to the world we started at, and use this new starting world as the start for our next action, thus meaning that each action is possible at the node it starts at in the sketch, and  making it possible that this sketch be a trace in the resulting model. The A* algorithm is used for action fitting since there a multiple ways to realize a given movement in the tension space, and we ideally want the optimal route and to avoid cyclical movements wherever possible. We do not provide pseudocode here since the A* algorithm is fairly well known and our modifications to it are not significant.

\subsection{Character Design}
\label{sec:chardesign}

\begin{figure}
\begin{subfigure}{\linewidth}
\centering
\begin{tabular}{|r|c|c|c|c|c|}
\hline
Char. & Theme & Prop. 1 & Prop. 2 & Prop. 3 & Prop. 4 \\
\hline
\multirow{3}{*}{Fanny} & Personal & F & T & T & F\\
& Family & F & T & F & T\\
& Society & F & F & T & F\\
\hline
\end{tabular}
\caption{The resulting worldviews for the sketch generated version of Fanny from \sas{}.}
\label{tab:generatedchar}
\end{subfigure}

\begin{subfigure}{\linewidth}
\centering
\begin{tabular}{|r|c|c|c|c|}
\hline
Bachelor & Prop. 1 & Prop. 2 & Prop. 3 & Prop. 4\\
\hline
Action 1 (Best) & F & T & T & (F)\\
Action 2 & F & (F) & (F) & (F)\\
Action 3 & F & T & (F) & T\\
\hline
\end{tabular}
\caption{The resulting values representing each bachelor. Technically the action postcondition. Parenthesis indicate that the action did not modify this value, thus it will remain the same as the actual world.}
\label{tab:generatedbach}
\end{subfigure}

\begin{subfigure}{\linewidth}
\centering
\begin{tabular}{|c|c|c|c|}
\hline
Prop. 1 & Prop. 2 & Prop. 3 & Prop. 4\\
\hline
T & F & F & F\\
\hline
\end{tabular}
\caption{The world from which all actions were generated, which we treat as the starting actual world.}
\label{tab:actualworld}
\end{subfigure}

\begin{subfigure}{\linewidth}
\centering
\begin{tabular}{|c|c|c|}
\hline
Pers. x Fam. & Fam. x Soc. & Soc. x Pers.\\
\hline
   \includegraphics[scale=0.12]{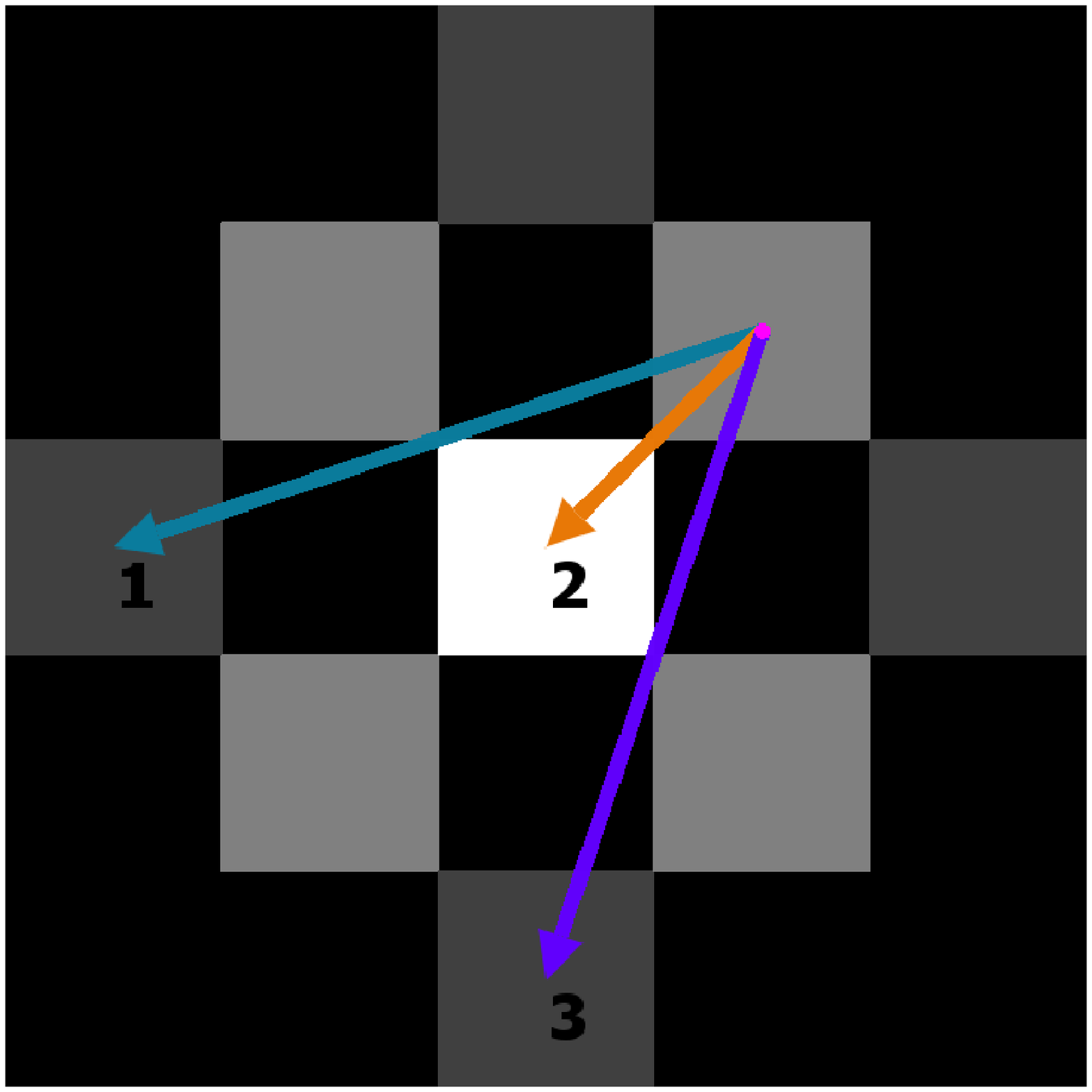}
 &  \includegraphics[scale=0.12]{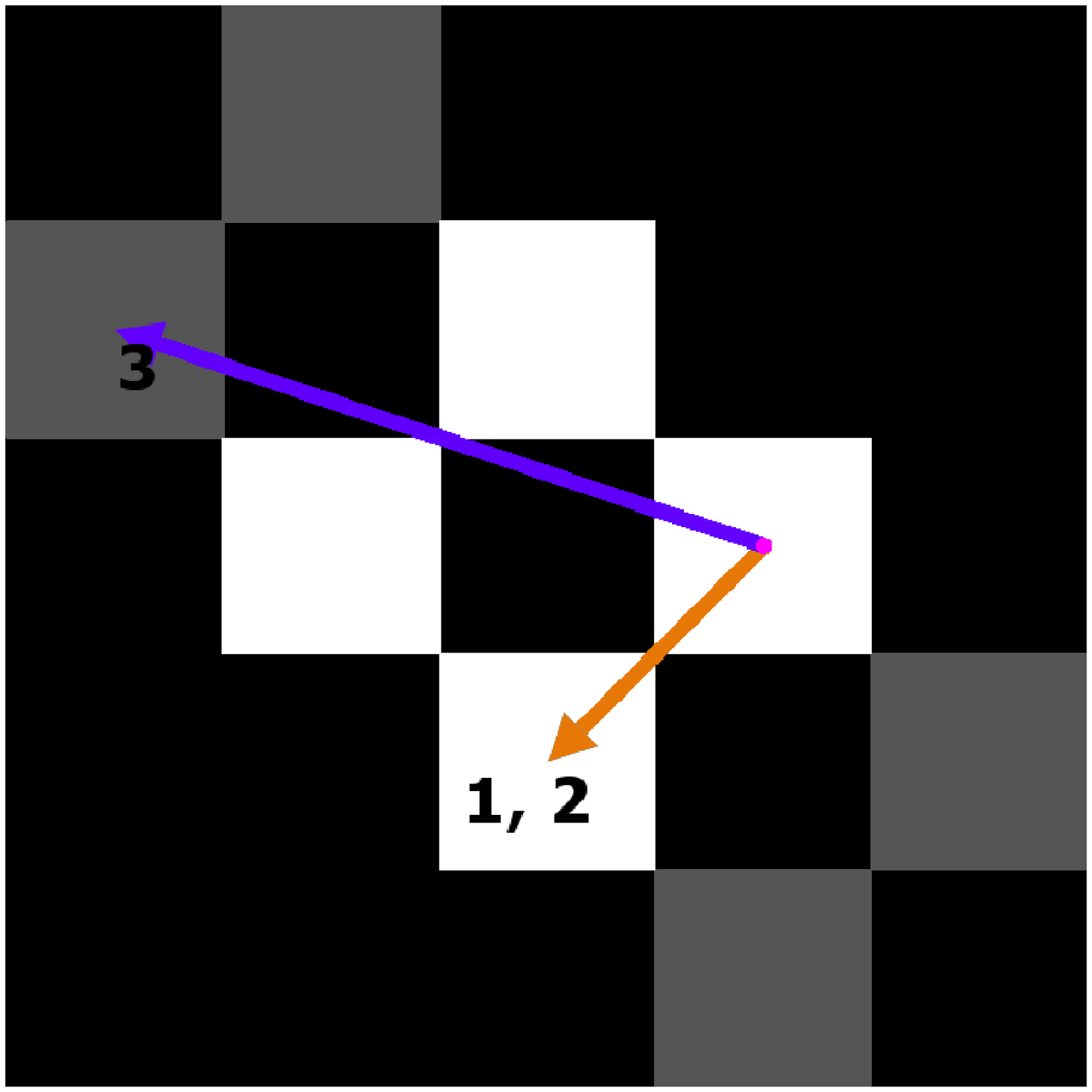}
 & \includegraphics[scale=0.12]{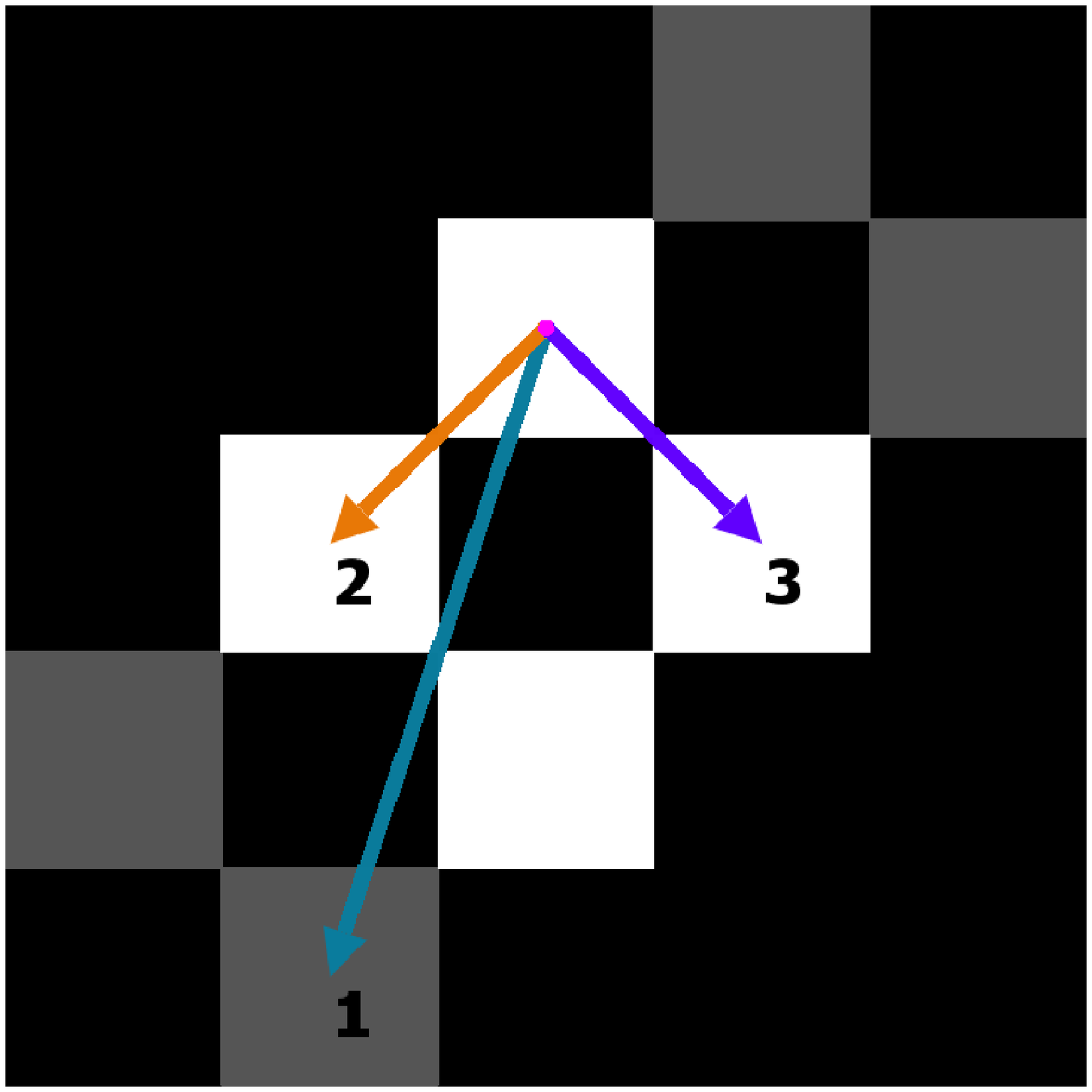} \\
\hline
\end{tabular}
\caption{The three main tension spaces and actions for the character, with different actions highlighted and labelled.}
\label{tab:generatedts}
\end{subfigure}

\caption{The generated character using the sketch approach. Shown is the character worldviews, actions (bachelors) and three main tension spaces. \sas.}
\label{tab:generatedsas}
\end{figure}

At this stage of development, we are interested in proving that the sketch approach is able to generate both tension spaces and actions that fit the intent of the author. Returning again to \sas{}, and looking at both the goals it had as well as the results we had through analysis, we wish to evaluate if it is possible to recreate the same character structure and actions using the sketch tool. We specifically look at the recreation of a single character, namely the character ``Fanny'', as each character is structurally the same. To use the sketching tool, the only parameters the author needs to provide is the number of characters, themes and propositions, and optionally label them if desired. For this experiment, we label the one character, Fanny, and the three themes as personal, society and family, since we aim to recreate her particular arrangement of tension spaces. We leave the four propositions unlabelled since we do not know yet how the values will be fit. The character sketching takes three steps, with five total sketches, which are shown in Figure~\ref{fig:sketchprocess}. \change{In Figure~\ref{fig:sketch1}, we use one sketch to recreate the personal/family tension space of Fanny, which roughly takes the shape of a diamond}. Notably, the sketch will only fit the first four movements of the sketch, but this still gives us the shape we want. Interestingly in these first two steps, there are technically only four movements, since there any inverse set of movements (eg. 4 and 8) can refer to the same proposition, so it is possible to fit eg. the first sketch with only the top half sketched. In Figure~\ref{fig:sketch2}, we complete the society worldview by sketching the society/family worldview, which takes the shape of a strong conflict space. Here, only the society worldview needs to be defined, so we define it in relation to one of the worldviews created in the first step. Lastly, we draw three sketches, one for each bachelor, recreating the movements from Figure~\ref{fig:analysistool}. We here also set the actual world to be the starting world of these actions, since this too must match the original character (this can be done automatically as part of the action sketching procedure). The sketch tool will always show the current state of the tension space during sketching as visual feedback, and when drawing actions each movement must lie somewhere within this tension space, since those are the only positions accessible between these worldviews.

The resulting character is shown in Figure~\ref{tab:generatedsas}. We can now attempt to assign each of the propositions to one of the propositions in \sas{}. Considering first the matched value, in \sas{} this value starts out as false, and each worldview wants it to be true, and further matching with any bachelor sets this value to be true. This pattern structurally exists, but in reverse, for proposition one. In the starting world, this value is true, and is false for all ideal worlds and false for each bachelor. Thus we could either reverse the values, or label the proposition as \textit{not matched} and the result is the same.  The remaining three propositions must represent some permutation of the wealth, faith and ambition propositions of the previous characters. In terms of these propositions, if we map proposition 2 to Faith, 3 to Wealth, and 4 to Ambition then the propositions perfectly match those of the original character.

Considering then the actions (bachelors). The bachelors, do not perfectly match the original values but provide the same functionality as the previous set of bachelors, with action 1 being the ideal character, action 3 being the red herring character, and action 2 being the remaining character. As noted in Figure\ref{tab:generatedts}, the tension spaces and movements even match those of \ref{fig:sasmovements}. Since the sketches only fit movements as needed, a number of the pre/post-conditions are set to \change{``don't care''}, since matching with that bachelor will not change the value of the actual world. This provides no difference in functionality, and is due to the optimal path found during the A* pathfinding fit of the movements.

This is a promising result, indicating that having a solid understanding of tension space and movement allows for the creation of a character with a strong set of design requirements. The character was additionally only created in around a minute with only five sketches, as opposed to the brute force approach of the original work. The tool additionally allows the option to ``refit'' the sketch and attempt to generate different results, which may be used during the tension phase sketching process to create various characters with matching structures. Considering re-implementing the full functionality of \sas{}, this is still likely outside the functionality of the tool, since with this approach if we made all three characters, it would similarly create three new bachelor actions each time, resulting in nine total actions. Exploring ways to achieve similar results, such as fitting tension spaces according to existing actions, is an interesting direction for future work.

There is an additional property to consider with the sketch tool. While we technically sketched two tension spaces to create this character, the third tension space occurs as a result of the first two and has no influence by the author. Further, there will always be one such space whenever dealing with an odd number of worldviews. In this example, the society/personal TS resulted from the personal worldview being created in the first step and the society worldview in the second. Although not an issue in this example, there are actually multiple shapes the resulting third tension space can take, based on the way the first two sketches are fit. Alternate ways to draw this character, such as drawing the personal/family TS followed instead by the society/personal TS can actually create another weak tension space for the family/society TS or a strong one, depending on the order the sketch is drawn in. In future, allowing access to changing the order in which sketches are fit would actually give more authorial control over the final space, but in general this will only be an issue when trying to perfectly fit a set of tension spaces, as we are doing here.

% Although this may cause problems when trying to perfectly fit a set of tension spaces as in this example, in general this is more an important feature to the ability of the tool to generate multiple characters from a set of sketches, and in future allowing access to refit sketches in different orders will allow better control over this final space.

%A further important result, is that formally one tension space will always occur as a result of previous sketches, which is the personal/society worldview in this case, which is defined based on the results of fitting a personal worldview in Figure~\ref{fig:sketch1} and a society worldview in Figure~\ref{fig:sketch2}. 

%Although not an issue in this example, in general due to this dependence, the final tension space may take multiple forms depending on the ordering of the previous sketches. This occurs since the worldview relations of the first two sketches defines the relations of the third. So if a proposition matches in the personal/family worldview, but differs in the family/society, we may conclude that the personal/society proposition will also differ. The ordering of these relations, however, is determined by the order of edges in the sketch, and it is possible that the same set of relations for the first two sketches will change the results of the third.

\section{Conclusions and Future Work}
\label{sec:conclusion}

Emergent narratives are a rich area of exploration into expressive means for creating and experiencing interactive narrative works with complex characters. Nonetheless the creation of EN works needs to be treated through a variety of lenses, not just in the authoring of content and systems, but also in the analysis and authoring support to help better understand the behaviour of the EN work during simulation. This analysis can apply to any number of intended results, but here we focused on analysing the foundational system according to its core properties of conflict and tension. We further showed as a proof-of-concept that it is possible to embrace the same approaches used for analysis to create new paths for authorship that focus first on the structural properties and behaviour of the EN work. These approaches lie outside any measure of narrative ``quality'', or qualitative judgement of the EN work, but rather are intended to assist in understanding the technical and formal properties of a highly complex systemic work. 

The foundational approach of this paper leaves each section open to further potential developments and properties. The PW system itself is still in active development, and already was significantly extended with dialogue and memory for \sas{}, and is further actively being developed with interpersonal and knowledge properties to support further emergent character behaviour. \change{The simplicity of our model, while still capable of interesting works, is still limited in several ways. For example, by only using one overall goal, characters are unable to adopt subgoals, or use any form of planning, and thus may avoid potential traces that lead to more ideal outcomes if one or more of those actions appear conflicting at the time. Intents may be explored by adding a lookahead, where a character simulates the outcomes of one or more steps of certain actions and selects the optimal one as a goal, however this approach is still in development.}

\change{Further without detailed models of belief manipulation and reasoning about other characters, we are incapable of modelling elements of deception and manipulation that are important to creating believable characters. This is currently ongoing work and is further leading to the development of a model of interpersonal character relationships. In this model, relationships are based upon \textit{memory}, and remembering which actions characters took and how the helped/hindered other character's overarching goals. In this sense, characters who help others may find the other characters taking actions that help them (even to the detriment of that character), and vice versa.}

\change{Another direction for improvement is better incorporation of the player. In this paper, we focused namely on characters, however there is a great potential in this analysis for looking at player behaviour. In this system, a player may take the role of an existing character, be an empty ``shell'' with no particular model, or even be able to define their own character. These would require different tools to better analyze the player's role in the narrative. One step towards this is the playthrough potential of the tool, which saves a playthrough of the work as a trace, that can be shown as a path on the tension space in the analysis tool. This can be used to display multiple player paths, either from existing playthroughs or even from simulated players, and give a better sense of the actual space available to the player, whereas we currently just look at the total possible space. The playthrough method can also be used to explore other aspects of the system, such as different action selection procedures and how they change the narratives which may occur.} 

Tension space analysis, in its reduction of an emergent narrative to an abstract space and movements, can be further expanded with coverage and reachability properties. The tool and in particular the sketch interface, are approached in this paper as proof-of-concept to show that such authoring capabilities are possible, but nonetheless identifying the role such a tool can play, be it as a declarative approach to authoring, or a more exploratory/brainstorming tool in the vein of the ``casual creator'', could further its development in different directions. 

\change{Scalability, as previously mentioned, is an ongoing concern, and at higher levels the tension space analysis will likely become more abstract, such as noting which spaces are strong/weak tension spaces, rather than rendering the complete tension space as is currently done. While a significant amount of work has been dedicated to optimization, scalability and ease of use must be considered with each new feature added to the system, and is certainly one of the limitations of this analysis.} \newchange{We plan on addressing scalability in a number of ways. First, we limit our analysis to well-defined components of narrative, in this case tension and conflict, rather than attempting to provide an exhaustive list of narrative properties. Second, limiting analysis towards a subset of the propositions in the world also provides a more fine-grained way and allows information to be processed sequentially. Third, we are working towards being able to automatically extract high level structures and patterns in the tension space that relate to certain narrative features~\cite{kybartas2018}. Authors could thus gain information about the patterns available in the space without the need to fully comprehend the visualization itself.}

The speed and ease of creating EN content at present, however remains a promising result which could be expanded with better control over the behaviour of generation to give more customizable results, such as controlling which propositions to generate for, or optimizing actions for certain movements. Through this expansion and development, we hope to discover new ways of both understanding and creating, new paths for emergent narrative creation.

\section{Acknowledgements}
This research was funded with the support of the Natural Sciences and Engineering Research Council of Canada (NSERC).

\bibliographystyle{IEEEtran}
\bibliography{references}

\begin{thebibliography}{10}
\providecommand{\url}[1]{#1}
\csname url@samestyle\endcsname
\providecommand{\newblock}{\relax}
\providecommand{\bibinfo}[2]{#2}
\providecommand{\BIBentrySTDinterwordspacing}{\spaceskip=0pt\relax}
\providecommand{\BIBentryALTinterwordstretchfactor}{4}
\providecommand{\BIBentryALTinterwordspacing}{\spaceskip=\fontdimen2\font plus
\BIBentryALTinterwordstretchfactor\fontdimen3\font minus
  \fontdimen4\font\relax}
\providecommand{\BIBforeignlanguage}[2]{{%
\expandafter\ifx\csname l@#1\endcsname\relax
\typeout{** WARNING: IEEEtran.bst: No hyphenation pattern has been}%
\typeout{** loaded for the language `#1'. Using the pattern for}%
\typeout{** the default language instead.}%
\else
\language=\csname l@#1\endcsname
\fi
#2}}
\providecommand{\BIBdecl}{\relax}
\BIBdecl

\bibitem{McCoy2014}
J.~McCoy, M.~Treanor, B.~Samuel, A.~A. Reed, M.~Mateas, and N.~Wardrip-Fruin,
  ``Social story worlds with {Comme il Faut},'' \emph{{IEEE} Transactions on
  Computational Intelligence and {AI} in Games}, Jun 2014.

\bibitem{aylett2005}
R.~S. Aylett, S.~Louchart, J.~Dias, A.~Paiva, and M.~Vala, ``Fearnot!
  {\textemdash} an experiment in emergent narrative,'' in \emph{Intelligent
  Virtual Agents}, T.~Panayiotopoulos, J.~Gratch, R.~Aylett, D.~Ballin,
  P.~Olivier, and T.~Rist, Eds.\hskip 1em plus 0.5em minus 0.4em\relax Berlin,
  Heidelberg: Springer, 2005, pp. 305--316.

\bibitem{adamsDwarfFortress2006}
{Bay 12 Games}, ``{Slaves to Armok: God of Blood Chapter II: Dwarf Fortress},''
  August 2006.

\bibitem{cavesofqud}
{Freehold Games}, ``{Caves of Qud},'' 2014.

\bibitem{paradoxCrusaderKings22012}
{Paradox Development Studio}, ``{Crusader Kings II},'' {Paradox Interactive},
  February 2012.

\bibitem{maxisSims}
{Maxis}, ``{The Sims},'' {Electronic Arts}, 2000.

\bibitem{jryanphd}
J.~Ryan, ``Curating simulated storyworlds,'' Ph.D. dissertation, University of
  California Santa Cruz, Dec. 2018.

\bibitem{eladhari2018}
M.~P. Eladhari, ``Re-tellings: The fourth layer of narrative as an instrument
  for critique,'' in \emph{Interactive Storytelling}, R.~Rouse, H.~Koenitz, and
  M.~Haahr, Eds.\hskip 1em plus 0.5em minus 0.4em\relax Cham: Springer
  International Publishing, 2018, pp. 65--78.

\bibitem{mateasInteractiveDramaPhD2002}
M.~Mateas, ``Interactive drama, art, and artificial intelligence,'' Ph.D.
  dissertation, Carnegie Mellon University, Pittsburgh, PA, December 2002.

\bibitem{garbeIceBound2014}
J.~Garbe, A.~A. Reed, M.~Dickinson, N.~Wardrip-Fruin, and M.~Mateas, ``Author
  assistance visualizations for ice-bound, a combinatorial narrative,'' in
  \emph{Proceedings of FDG 2014 - Ninth International Conference on the
  Foundations of Digital Games}, 2014.

\bibitem{smelikDeclarativeModeling2011}
R.~M. Smelik, T.~Tutenel, K.~J. De~Kraker, and R.~Bidarra, ``A declarative
  approach to procedural modeling of virtual worlds,'' \emph{Computers and
  Graphics}, vol.~35, no.~2, pp. 352--363, Apr. 2011.

\bibitem{liapis2013sketchbook}
A.~Liapis, G.~N. Yannakakis, and J.~Togelius, ``Sentient sketchbook:
  Computer-aided game level authoring,'' in \emph{Proceedings of the 8th
  Conference on the Foundations of Digital Games}, 2013, pp. 213--220.

\bibitem{compton-iccc2015}
K.~Compton and M.~Mateas, ``Casual creators,'' in \emph{Proceedings of the
  Sixth International Conference on Computational Creativity (ICCC 2015)},
  H.~Toivonen, S.~Colton, M.~Cook, and D.~Ventura, Eds.\hskip 1em plus 0.5em
  minus 0.4em\relax Park City, Utah: Brigham Young University, Jun.--Jul. 2015,
  pp. 228--235.

\bibitem{koenitzchallenges2019}
H.~Koenitz and M.~P. Eladhari, ``Challenges of idn research and teaching,'' in
  \emph{Interactive Storytelling}, R.~E. Cardona-Rivera, A.~Sullivan, and R.~M.
  Young, Eds.\hskip 1em plus 0.5em minus 0.4em\relax Cham: Springer
  International Publishing, 2019, pp. 26--39.

\bibitem{kybartas_2017}
B.~Kybartas, C.~Verbrugge, and J.~Lessard, ``Subject and subjectivity: A
  conversational game using possible worlds,'' in \emph{Interactive
  Storytelling}.\hskip 1em plus 0.5em minus 0.4em\relax Springer International
  Publishing, 2017.

\bibitem{subjectandsubjectivity}
\BIBentryALTinterwordspacing
B.~Kybartas, ``Subject and subjectivity,'' 2017. [Online]. Available:
  \url{https://tineola.itch.io/subject-and-subjectivity}
\BIBentrySTDinterwordspacing

\bibitem{kybartas2018}
B.~Kybartas, C.~Verbrugge, and J.~Lessard, ``Expressive range analysis of a
  possible worlds driven emergent narrative system,'' in \emph{Interactive
  Storytelling}, R.~Rouse, H.~Koenitz, and M.~Haahr, Eds.\hskip 1em plus 0.5em
  minus 0.4em\relax Cham: Springer International Publishing, 2018, pp.
  473--477.

\bibitem{Pavel1975}
T.~G. Pavel, ``{"Possible Worlds"} in literary semantics,'' \emph{The Journal
  of Aesthetics and Art Criticism}, vol.~34, no.~2, p. 165, 1975.

\bibitem{dolozel2000}
L.~Dole{\v{z}}el, \emph{Heterocosmica: Fiction and Possible Worlds}, ser.
  Parallax: Re-visions of Culture and Society.\hskip 1em plus 0.5em minus
  0.4em\relax Johns Hopkins University Press, 2000.

\bibitem{Ryan1991}
M.-L. Ryan, \emph{Possible Worlds, Artificial Intelligence, and Narrative
  Theory}.\hskip 1em plus 0.5em minus 0.4em\relax Bloomington, IN, USA: Indiana
  University Press, 1991.

\bibitem{evansVersu2014}
R.~Evans and E.~Short, ``{Versu} --- a simulationist storytelling system,''
  \emph{IEEE Transactions on Computational Intelligence and AI in Games},
  vol.~6, no.~2, pp. 113--130, June 2014.

\bibitem{Lessard2016}
J.~Lessard and D.~Arsenault, ``The character as subjective interface,'' in
  \emph{Interactive Storytelling {\textemdash} 9th International Conference on
  Interactive Digital Storytelling}, F.~Nack and A.~S. Gordon, Eds.\hskip 1em
  plus 0.5em minus 0.4em\relax Los Angeles, CA, USA: Springer International
  Publishing, Nov. 2016, pp. 317--324.

\bibitem{harrellbook2013}
D.~F. Harrell, \emph{Phantasmal Media - An Approach to Imagination,
  Computation, and Expression}.\hskip 1em plus 0.5em minus 0.4em\relax
  Cambridge, Massachusetts: MIT Press, 2013.

\bibitem{sgourosQuantumConceptsNarrative2015}
N.~M. Sgouros, ``Embedding and implementation of quantum computational concepts
  in digital narratives,'' in \emph{Entertainment Computing - {ICEC} 2015},
  K.~Chorianopoulos, M.~Divitini, J.~Baalsrud~Hauge, L.~Jaccheri, and
  R.~Malaka, Eds.\hskip 1em plus 0.5em minus 0.4em\relax Cham: Springer
  International Publishing, 2015, pp. 140--154.

\bibitem{smithExpressiveRange2010}
G.~Smith and J.~Whitehead, ``Analyzing the expressive range of a level
  generator,'' in \emph{Proceedings of the 2010 Workshop on Procedural Content
  Generation in Games}.\hskip 1em plus 0.5em minus 0.4em\relax ACM, 2010.

\bibitem{wareConflict2014}
S.~G. Ware, R.~M. Young, B.~Harrison, and D.~L. Roberts, ``A computational
  model of plan-based narrative conflict at the fabula level,'' \emph{{IEEE}
  Transactions on Computational Intelligence and {AI} in Games}, vol.~6, no.~3,
  pp. 271--288, sep 2014.

\bibitem{szilasDrama2017}
N.~Szilas, ``Modeling and representing dramatic situations as paradoxical
  structures,'' \emph{Digital Scholarship in the Humanities}, vol.~32, no.~2,
  pp. 403--422, feb 2016.

\bibitem{7439785}
B.~Kybartas and R.~Bidarra, ``A survey on story generation techniques for
  authoring computational narratives,'' \emph{IEEE Transactions on
  Computational Intelligence and AI in Games}, vol.~9, no.~3, pp. 239--253,
  Sept 2017.

\bibitem{aylett1999}
R.~Aylett, ``Narrative in virtual environments {\textemdash} towards emergent
  narrative,'' in \emph{AAAI Symposium on Narrative Intelligence}, 01 1999, pp.
  83--86.

\bibitem{swartjesLateCommitment2008}
I.~Swartjes, E.~Kruizinga, and M.~Theune, ``\BIBforeignlanguage{English}{Let's
  pretend {I} had a sword},'' in \emph{\BIBforeignlanguage{English}{Interactive
  Storytelling}}, ser. Lecture Notes in Computer Science, U.~Spierling and
  N.~Szilas, Eds.\hskip 1em plus 0.5em minus 0.4em\relax Springer Berlin
  Heidelberg, 2008, vol. 5334, pp. 264--267.

\bibitem{Fuller2010}
D.~Fuller and B.~Magerko, ``Shared mental models in improvisational
  performance,'' in \emph{Proceedings of the Intelligent Narrative Technologies
  {III} Workshop on - {INT}3 '10}.\hskip 1em plus 0.5em minus 0.4em\relax {ACM}
  Press, 2010.

\bibitem{nothingfordinner2014}
\BIBentryALTinterwordspacing
N.~Szilas, J.~Dumas, U.~Richle, T.~Boggini, and N.~Habonneau, ``Nothing for
  dinner,'' 2014. [Online]. Available:
  \url{http://nothingfordinner.org/portal/}
\BIBentrySTDinterwordspacing

\bibitem{szilasIDTension2003fixed}
N.~Szilas, ``Idtension: A narrative engine for interactive drama,'' in
  \emph{Proceedings of the Technologies for Interactive Digital Storytelling
  and Entertainment (TIDSE) Conference}, Darmstadt, Germany, Mar. 2003.

\bibitem{AIIDE1715844}
B.~Thorne and R.~M. Young, ``Generating stories that include failed actions by
  modeling false character beliefs,'' in \emph{Intelligent Narrative
  Technologies}.\hskip 1em plus 0.5em minus 0.4em\relax The AAAI Press, 2017.

\bibitem{Shirvani_Ware_Farrell_2017}
A.~Shirvani, S.~G. Ware, and R.~Farrell, ``A possible worlds model of belief
  for state-space narrative planning,'' in \emph{Proceedings of the Thirteenth
  AAAI Conference on Artificial Intelligence and Interactive Digital
  Entertainment}, 2017.

\bibitem{Si:2005:TUM:1082473.1082477}
M.~Si, S.~C. Marsella, and D.~V. Pynadath, ``Thespian: Using multi-agent
  fitting to craft interactive drama,'' in \emph{Proceedings of the Fourth
  International Joint Conference on Autonomous Agents and Multiagent Systems},
  ser. AAMAS '05.\hskip 1em plus 0.5em minus 0.4em\relax New York, NY, USA:
  ACM, 2005, pp. 21--28.

\bibitem{alfonso2015emotional}
B.~Alfonso, D.~V. Pynadath, M.~Lhommet, and S.~Marsella, ``Emotional perception
  for updating agents' beliefs,'' in \emph{Affective Computing and Intelligent
  Interaction (ACII), 2015 International Conference on}.\hskip 1em plus 0.5em
  minus 0.4em\relax IEEE, 2015, pp. 201--207.

\bibitem{El-Nasr:2004:UAS:1067343.1067356}
M.~S. El-Nasr, ``A user-centric adaptive story architecture: Borrowing from
  acting theories,'' in \emph{Proceedings of the 2004 ACM SIGCHI International
  Conference on Advances in Computer Entertainment Technology}, ser. ACE
  '04.\hskip 1em plus 0.5em minus 0.4em\relax New York, NY, USA: ACM, 2004, pp.
  109--116.

\bibitem{egerBeliefManipulation}
M.~Eger and C.~Martens, ``Practical specification of belief manipulation in
  games,'' in \emph{Proceedings of the Thirteenth AAAI Conference on Artificial
  Intelligence and Interactive Digital Entertainment}, 2017.

\bibitem{robertson2019}
J.~{Robertson} and R.~M. {Young}, ``Perceptual experience management,''
  \emph{IEEE Transactions on Games}, vol.~11, no.~1, pp. 15--24, March 2019.

\bibitem{Harrell2018}
D.~F. Harrell, P.~Ortiz, P.~Downs, M.~Wagoner, E.~Carre, and A.~Wang,
  ``Chimeria:grayscale: an interactive narrative for provoking critical
  reflection on gender discrimination,'' \emph{Materialities of Literature},
  vol.~6, no.~2, pp. 217--221, 2018.

\bibitem{harrell2007b}
D.~F. Harrell, \emph{Second Person: Role-Playing and Story in Games and
  Playable Media}.\hskip 1em plus 0.5em minus 0.4em\relax Cambridge,
  Massachusetts: MIT Press Ltd, 2007, ch. GRIOT's Tales of Haints and Seraphs:
  A Computational Narrative Generation System, pp. 177--182.

\bibitem{peinado2008}
F.~Peinado, M.~Cavazza, and D.~Pizzi, ``Revisiting character-based affective
  storytelling under a narrative bdi framework,'' in \emph{Interactive
  Storytelling}, U.~Spierling and N.~Szilas, Eds.\hskip 1em plus 0.5em minus
  0.4em\relax Berlin, Heidelberg: Springer, 2008, pp. 83--88.

\bibitem{berov2017}
L.~Berov, ``Steering plot through personality and affect: An extended {BDI}
  model of fictional characters,'' in \emph{KI 2017: Advances in Artificial
  Intelligence}, G.~Kern-Isberner, J.~F{\"u}rnkranz, and M.~Thimm, Eds.\hskip
  1em plus 0.5em minus 0.4em\relax Cham: Springer International Publishing,
  2017, pp. 293--299.

\bibitem{kybartasReGEN2014}
B.~Kybartas and C.~Verbrugge, ``Analysis of {ReGEN} as a graph-rewriting system
  for quest generation,'' \emph{{IEEE} Transactions on Computational
  Intelligence and {AI} in Games}, vol.~6, no.~2, pp. 228--242, jun 2014.

\bibitem{baesurprise2014}
B.-C. Bae and R.~Young, ``A computational model of narrative generation for
  surprise arousal,'' \emph{Computational Intelligence and AI in Games, IEEE
  Transactions on}, vol.~6, no.~2, pp. 131--143, June 2014.

\bibitem{szilas2018}
N.~Szilas, S.~Estupi{\~{n}}{\'a}n, and U.~Richle, ``Automatic detection of
  conflicts in complex narrative structures,'' in \emph{Interactive
  Storytelling}, R.~Rouse, H.~Koenitz, and M.~Haahr, Eds.\hskip 1em plus 0.5em
  minus 0.4em\relax Cham: Springer International Publishing, 2018, pp.
  415--427.

\bibitem{AIIDE1818103}
N.~Partlan, E.~Carstensdottir, S.~Snodgrass, E.~Kleinman, G.~Smith,
  C.~Harteveld, and M.~S. El-Nasr, ``Exploratory automated analysis of
  structural features of interactive narrative,'' 2018.

\bibitem{ryanPossibleWorldsNarrative1991}
M.-L. Ryan, \emph{Possible Worlds, Artificial Intelligence, and Narrative
  Theory}.\hskip 1em plus 0.5em minus 0.4em\relax Bloomington, IN, USA: Indiana
  University Press, 1991.

\bibitem{berovTellability2017}
L.~Berov, ``Towards a computational measure of plot tellability,'' in
  \emph{Intelligent Narrative Technologies}.\hskip 1em plus 0.5em minus
  0.4em\relax The AAAI Press, 2017.

\bibitem{siProactiveAuthoring2007}
M.~Si, S.~C. Marsella, and D.~V. Pynadath, ``Proactive authoring for
  interactive drama: An author's assistant,'' in \emph{Intelligent Virtual
  Agents}, C.~Pelachaud, J.-C. Martin, E.~Andr{\'e}, G.~Chollet, K.~Karpouzis,
  and D.~Pel{\'e}, Eds.\hskip 1em plus 0.5em minus 0.4em\relax Berlin,
  Heidelberg: Springer, 2007, pp. 225--237.

\end{thebibliography}

\end{document}